\documentclass[
  aip,cha,            
  preprint,           
  amsmath,amssymb,
  longbibliography
]{revtex4-2}

\usepackage{graphicx}   
\usepackage{dcolumn}    
\usepackage{bm}         
\usepackage{xcolor}
\usepackage[normalem]{ulem} 
\usepackage{booktabs}
\usepackage{tabularx}
\usepackage{adjustbox}
\usepackage{array}
\usepackage{url}
\usepackage[caption=false]{subfig}

\usepackage{tikz}
\usetikzlibrary{shapes, positioning, arrows.meta, fit, backgrounds}

\tikzset{
  every node/.style={font=\sffamily, minimum height=0.8cm, inner sep=4pt},
  box/.style={draw, fill=gray!10, rounded corners=2pt, line width=0.8pt, minimum width=2.0cm},
  conv/.style={box, fill=cyan!10},
  op/.style={draw, circle, fill=gray!10, line width=0.8pt, minimum size=0.8cm},
  arrow/.style={->, line width=0.8pt, >={Latex[round,open]}}
}

\newcommand{\alpgt}{\alpha^\mathrm{gt}}

\begin{document}

\title[DCIts for multivariate time series]{Interpretable deep convolutional model for nonlinear multivariate time series in complex systems}

\author{Domjan Bari\'c}
\author{Davor Horvati\'c}
\email{davorh@phy.hr}
\thanks{Corresponding author.}
\affiliation{Department of Physics, Faculty of Science, University of Zagreb, Bijeni\v{c}ka cesta 32, 10000 Zagreb, Croatia}

\date{\today}

\begin{abstract}
We introduce the Deep Convolutional Interpreter for Time Series (DCIts), a deep learning architecture for nonlinear multivariate time series that provides sample-specific, locally interpretable descriptions of the underlying interaction structure. Unlike standard black-box forecasters, DCIts learns a time- and lag-dependent transition tensor explicitly factorized into two components: a \emph{Focuser}, which selects relevant source series and time lags via a sparse masking mechanism, and a \emph{Modeler}, which assigns signed coefficients to these selected interactions. This decomposition yields a local lag-adjacency structure and signed source--lag contributions for every forecast instance, enabling direct inspection of effective connectivity; when higher-order branches are activated, the same framework yields order-resolved elementwise polynomial contributions. Architecturally, DCIts uses a diverse bank of convolutional filters to capture temporal and cross-variable dependencies, which are mapped through a bottleneck network to the transition tensor. On controlled benchmark datasets with known interaction structure, we demonstrate that DCIts achieves competitive forecasting error relative to a strong interpretable baseline while recovering stable, signed, lag-resolved interaction patterns. The framework thus prioritizes intrinsic interpretability, using forecasting accuracy as a faithfulness constraint rather than the sole objective.
\end{abstract}

\maketitle

\begin{quotation}
Nonlinear interactions among components are a defining feature of complex dynamical systems and are frequently observed through multivariate time series. Yet, many high-performing deep learning forecasters remain difficult to interpret with respect to which variables influence a prediction, at which lags, and with what sign and magnitude. We present DCIts, a deep convolutional architecture that learns a time- and lag-dependent transition tensor and factorizes it into a sparse \emph{Focuser} mask and a signed \emph{Modeler} tensor, yielding a per-sample, lag-resolved interaction pattern for every forecast. Across benchmarking datasets with known ground truth, DCIts achieves forecasting accuracy competitive with the selected interpretable baseline while providing stable local explanations of coupling and effective connectivity.
\end{quotation}

\section{Introduction}
\label{intro}

Multivariate time series provide a primary empirical window into the dynamics of complex systems, from coupled oscillators, geophysical and climate subsystems, and ecological networks to physiological and neural processes. In such settings, each component series both influences and is influenced by others through nonlinear and time-lagged interactions that may vary across operating regimes. Beyond prediction alone, identifying these interactions is often central to the scientific objective: it supports mechanistic interpretation, effective connectivity analysis, instability diagnosis, and the anticipation of regime shifts. In this paper, forecasting is used primarily as a training objective that forces learned explanations to remain faithful to the predictive mechanism, rather than as the end goal.

At the same time, forecasting high-dimensional time series remains challenging because relevant dependencies may be weak, delayed, nonlinear, nonstationary, and confounded by common drivers~\cite{lim2021time,Rungeetal19}. Recent deep learning architectures have demonstrated strong empirical performance for multivariate forecasting, frequently surpassing classical statistical baselines in settings with complex seasonalities, long-range temporal structure, and cross-series dependencies~\cite{lim2021time,benidis2022deep,chen2023tsmixer}. However, these gains are typically achieved with high-capacity models---including recurrent, convolutional, and attention-based architectures---whose internal representations are not directly interpretable in terms of source variables, lags, signs, and interaction strengths~\cite{baric2021,fauvel2020performance,rudin2019stop}. This limits their usefulness in settings where forecasts must be inspected, trusted, or translated into domain-relevant explanations, especially in safety- and accountability-critical applications such as healthcare~\cite{rudin2019stop,tonekaboni2020went}.

This tension between predictive accuracy and interpretability has motivated a rapidly growing body of work on explainable forecasting models for multivariate time series. A useful perspective is provided by Bari\'c et al.\ \cite{baric2021}, who review interpretable approaches and propose benchmarking datasets and evaluation criteria. Within the Performance--Explainability Framework \cite{fauvel2020performance}, models are assessed not only by predictive performance but also by criteria such as comprehensibility, granularity, faithfulness, and end-user considerations. A recurrent conclusion is that many methods offer either strong forecasting performance with limited transparency or intuitive explanations that are insufficiently faithful to the underlying predictive mechanism \cite{baric2021,fauvel2020performance}. For example, architectures such as IMV-LSTM \cite{guo2019exploring} explicitly structure hidden states to isolate feature-specific information, improving interpretability, but the broader challenge remains: how to obtain explanations that are both local (forecast-specific) and faithful while retaining competitive forecasting accuracy across datasets.

A prominent family of approaches seeks interpretability via post-hoc attribution or model-agnostic explanation methods, including SHAP~\cite{lundberg2017unified} and perturbation-based analyses~\cite{perturbation}, as well as by inspecting learned attention weights~\cite{Tryambak2020,Pham2023,Wu2023,9308570,Feng2022,GANDIN2021103876}. This includes dual-stage attention models such as DA-RNN~\cite{DualStage}, in which input attention is used to weight driving series and temporal attention is used to weight encoder states. These tools can be valuable for exploratory analysis, but they can also yield explanations that are sensitive to correlations in the inputs, architecture-dependent, or difficult to interpret mechanistically. More recently, counterfactual and language-based explanation protocols have been proposed specifically for forecasting tasks~\cite{wang_counterfactual_2023,xforecast2024}. Such language-based methods can improve the accessibility of explanations, but they are generally explanation protocols placed around or after a forecasting model, rather than architectures that directly expose a signed local transition law. In the specific case of attention, there is an active debate over when attention distributions should be treated as explanations~\cite{attentionNot,attentionNotNot}. Attention weights can provide useful relevance information and, in carefully designed forecasting architectures, can support interpretable summaries of input or temporal importance~\cite{DualStage,guo2019exploring,baric2021}. However, attention weights are typically relevance scores over inputs, time steps, or hidden states; they are not generally signed local transition coefficients and therefore do not, by themselves, provide an explicit representation of \emph{who influences whom}, \emph{at which lags}, and \emph{with what sign and functional form}. This coefficient-level representation is the specific interpretability target of DCIts.

An alternative strategy is to design forecasting architectures that are interpretable \emph{by construction}. Notable examples include N-BEATS \cite{Oreshkin2019}, which decomposes univariate forecasts into structured components such as trend and seasonality, and models that expose probabilistic parameters or latent factors for interpretability \cite{Ozyegen2022,Rasul2020}. A complementary line of work aims to enforce interpretability in otherwise opaque forecasting backbones (e.g., Transformers) by incorporating explicit interpretable bottlenecks \cite{vansprang2024conceptbottleneck}. Frequency- or scale-aware decompositions can also improve interpretability for specific tasks \cite{assaf2019explainable,Wang2018}. While these approaches represent important progress, they typically impose structural constraints (e.g., univariate settings, predefined component families, distributional assumptions, or task-specific decompositions) and do not directly provide a forecast-local, signed, lag-resolved description of multivariate interactions.

Relatedly, there is a substantial literature on causal discovery and dependency analysis in time series, including approaches based on conditional independence testing and graph discovery \cite{Runge2018,Rungeetal19,Runge_2019}. More recently, ensemble methods combined with feature-importance analyses have been proposed to infer lagged causal relationships \cite{castro2023time}. These methods can be effective for detecting whether dependencies exist, but they often do not yield an explicit, interpretable representation of \emph{how} a source series at a given lag contributes to a specific forecast (e.g., sign, strength, and nonlinear order), and they may require preprocessing choices such as detrending that can influence conclusions \cite{castro2023time}. For many applications in complex systems, what is still missing is a modelling framework that jointly supports competitive nonlinear multivariate forecasting and a \emph{forecast-specific} and \emph{low-dimensional} representation of the local interaction structure.

In this work, we introduce an interpretable deep learning model for multivariate time-series forecasting designed to address this specific coefficient-level interpretability gap. The model learns an interaction tensor that is \emph{explicitly factorized} into a \emph{Focuser}, which selects relevant source series and time lags through a sparse mask, and a \emph{Modeler}, which assigns signed coefficients to the selected interactions. This yields, for each target series and each forecast instance, a local lag-adjacency pattern together with signed source--lag contributions; when higher-order branches are activated, these contributions can be resolved by polynomial order. As a result, the model provides direct, inspection-ready information about coupling structure and effective connectivity while maintaining a forecasting objective.

Beyond interpretability at the level of individual forecasts, the architecture is designed to reduce manual supervision in model specification. In particular, it supports principled selection of the window length (to capture relevant lagged interactions within the smallest sufficient history) and identification of an appropriate model order (balancing expressiveness and complexity when higher-order terms are included). Through experiments on benchmark datasets \cite{baric2021}, we demonstrate that the model can recover salient series, lags, and interaction orders, providing faithful and granular explanations that complement its predictive performance.

The remainder of the paper is organized as follows: we first define the proposed architecture and its factorized interaction representation, then present the experimental protocols and datasets, and finally analyze forecasting performance, along with the induced lag-adjacency and contribution patterns, to illustrate the resulting interpretability.

\section{Deep Convolutional Interpreter for time series}\label{DCIts}

We introduce a novel architecture designed for modeling multivariate time series (MTS). An MTS system consists of $N$ time series, each of length $M$, collectively represented by $\bm{X} = {\bm{X}_1, \bm{X}_2, \ldots , \bm{X}_N}$. The $n$-th time series in this system is denoted by $\bm{X}_n = (X_{n,1} , X_{n,2} , \ldots , X_{n,M} )$, where $X_{n,t} \in \mathbb{R}$ signifies the value of the $n$-th time series at time $t$. Here, $n = 1, 2, \ldots, N \in \mathbb{N}$ indexes the series, and $t = 1, 2, \ldots, M \in \mathbb{N}$ indexes time points within each series.
We aim to analyze and decipher the interactions among various time series and their respective lags. To achieve this, we have developed a deep convolutional interpreter for time series (DCIts). This architecture is designed to approximate the transition tensor $\bm{\alpha}_t$, which describes how the time series evolves between consecutive time points, utilizing the $L$ lags in the MTS window $\bm{Q}_{t}$.

DCIts operates on sliding windows of length $L$. We define a fixed-size window, $L$, with $L < M$, to segment MTS into overlapping intervals, and construct a sequence of windows, denoted as $\bm{Q}_{t}$. Each window $\bm{Q}_{t}$ is defined as:
\begin{equation}
\bm{Q}_{t} = (\bm{X}_{t}, \bm{X}_{t-1}, \bm{X}_{t-2}, \ldots, \bm{X}_{t-L+1}), \quad \text{for} \quad t = L, L+1, \ldots, M.
\end{equation}

We collect windowed samples in a tensor $\bm{Q}\in\mathbb{R}^{(M-L+1)\times N\times L}$, where $\bm{Q}_t\in\mathbb{R}^{N\times L}$ denotes the $t$-th windowed sample.\footnote{To prevent leakage with rolling windows, we first split the original series $\bm{X}$ into contiguous train/validation/test segments and then construct windowed samples $\bm{Q}_t$ \emph{within} each segment; windows that would cross a split boundary are not formed. Only the training windows are shuffled; validation and test windows preserve temporal order.}

Let $\bm{Q}_t \in \mathbb{R}^{N \times L}$ be the window of length $L$ capturing the most recent $L$ observations from each of the $N$ time series, and let $\bm{\alpha}_t \in \mathbb{R}^{N \times N \times L}$ be a tensor whose entry $(\alpha_t)_{n,i,\ell}$ quantifies the contribution of source series $i$ at lag $\ell$ to the prediction of target series $n$.

We write the one-step-ahead transition explicitly as
\begin{equation}\label{eq:transition_diamond}
\hat X_{n,t+1}
=
\sum_{i=1}^{N}\sum_{\ell=1}^{L}
(\alpha_t)_{n,i,\ell}\,(Q_t)_{i,\ell},
\qquad n=1,\ldots,N .
\end{equation}
Here, $(\alpha_t)_{n,i,\ell}$ is the local coefficient multiplying source series $i$ at lag $\ell$ in the prediction of target series $n$. For compact notation, we denote the corresponding source--lag contribution tensor by $\bm{\alpha}_t\diamondsuit\bm{Q}_t$, with entries
\begin{equation}\label{eq:hadamard_revised}
\bigl(\bm{\alpha}_t\diamondsuit\bm{Q}_t\bigr)_{n,i,\ell}
=
(\alpha_t)_{n,i,\ell}(Q_t)_{i,\ell}.
\end{equation}
Thus, $\diamondsuit$ denotes only this broadcasted elementwise multiplication between the transition tensor and the input window; it introduces no additional modelling operation beyond the explicit summation in Eq.~\eqref{eq:transition_diamond}. The tensor $\bm{\alpha}_t\diamondsuit\bm{Q}_t$ lies in $\mathbb{R}^{N\times N\times L}$, and summing it over source series and lags produces the one-step forecast $\hat X_{n,t+1}$.

For clarity, Table~\ref{tab:shape_summary} summarizes the main tensors and their dimensions.
\begin{table}[!htbp]
\centering
{
\begin{tabular}{lll}
\toprule
Symbol & Shape & Meaning \\
\midrule
$\bm{X}$ & $\mathbb{R}^{N\times M}$ & full multivariate time series \\
$\bm{Q}_t$ & $\mathbb{R}^{N\times L}$ & input window at time $t$ \\
$\bm{Z}_t$ & $\mathbb{R}^{N\times N\times L}$ & Focuser pre-activation tensor \\
$\bm{F}_t$ & $(0,1)^{N\times N\times L}$ & Focuser mask \\
$\bm{C}_t$ & $\mathbb{R}^{N\times N\times L}$ & signed Modeler coefficient tensor \\
$\bm{\alpha}_t=\bm{C}_t\circ\bm{F}_t$ & $\mathbb{R}^{N\times N\times L}$ & transition tensor \\
$\bm{\alpha}_t\diamondsuit\bm{Q}_t$ & $\mathbb{R}^{N\times N\times L}$ & source--lag contribution tensor \\
$\hat{\bm{X}}_{t+1}$ & $\mathbb{R}^{N}$ & one-step forecast vector \\
\bottomrule
\end{tabular}
}
\caption{Notation and tensor shapes used in the DCIts forward pass. Here $N$ is the number of time series, $M$ is the full series length, and $L$ is the input-window length.}
\label{tab:shape_summary}
\end{table}
The forward pass can therefore be summarized as follows. The Focuser maps the input window $\bm{Q}_t$ to a pre-activation tensor $\bm{Z}_t$ and then to a mask $\bm{F}_t=\sigma_T(\bm{Z}_t)$. The Modeler receives the masked input $\bm{F}_t\diamondsuit\bm{Q}_t$ and produces signed coefficients $\bm{C}_t$. Their Hadamard product gives the transition tensor $\bm{\alpha}_t=\bm{C}_t\circ\bm{F}_t$, and the forecast is obtained by summing the contribution tensor $\bm{\alpha}_t\diamondsuit\bm{Q}_t$ over source series and lags.

Moreover, this tensor allows an interpretation of the MTS system's internal dynamics. Specifically, it enables deciphering of the underlying equations that generate the MTS. It is important to distinguish $\bm{\alpha}_t$ from an attention tensor. Attention coefficients are typically non-negative relevance weights, often normalized over inputs or hidden states, and therefore indicate which parts of the representation are emphasized by the model. By contrast, $\bm{\alpha}_t$ is a signed, lag-resolved transition-coefficient tensor. Its entries are not constrained to sum to one; instead, each entry directly multiplies a source-series value at a specific lag in Eq.~\eqref{eq:transition_diamond}. Thus, $\alpha_{n,i,\ell}$ is interpretable as a local coefficient of the contribution of source series $i$ at lag $\ell$ to target series $n$, including its sign and magnitude.

To realize its objectives, DCIts employs a two-step predictive strategy. The first component, the \textit{Focuser}, identifies significant lags and time series - effectively separating the signal from noise.  Following this, the \textit{Modeler} phase leverages the insights from the Focuser, alongside the original input, to deduce the final coefficients for the transition tensor $\bm{\alpha}_t$. The final result is calculated using the input, Focuser, and Modeler outputs. The architecture is described in detail below and summarized schematically in Fig.~\ref{ARH} after the relevant quantities have been defined.

\subsection{Focuser and Modeler}\label{focuser-modelar}

Both the Focuser and Modeler use the same overall architecture (shown in Figure \ref{ARH}), the only difference is in the final layer. The motivation for this two-branch construction is to separate two conceptually distinct tasks. The Focuser estimates which source--lag entries are relevant for the current forecast window, while the Modeler estimates the signed magnitude of their contribution. This separation is analogous to the distinction between support selection and coefficient estimation in sparse modelling: the Focuser provides a soft, sample-specific support mask, whereas the Modeler provides the corresponding signed local coefficients. The final object of interpretation is therefore not the Focuser mask alone, but the transition tensor $\bm{\alpha}_t=\bm{C}_t\circ\bm{F}_t$. In both Focuser and Modeler, we first apply a set of convolutional kernels with varying shapes. In our benchmarks, the convolutional coefficient-generating architecture was sufficient, so we do not add a separate attention block. This choice keeps the explanatory object tied directly to the prediction equation: attention weights are typically relevance scores over inputs or hidden states, whereas DCIts exposes signed transition coefficients linking target series, source series, lags, and magnitudes.

The convolutional bank contains kernels spanning complementary parts of the input window: a global $(N\times L)$ kernel, per-series temporal kernels $(1\times L)$, cross-series instantaneous kernels $(N\times 1)$, cross-variable short-lag kernels $(N\times 3)$ and $(N\times 5)$, and per-series short-lag kernels $(1\times 3)$ and $(1\times 5)$. Kernels with height $N$ span all source series, whereas kernels with height $1$ act per series. Thus, the architecture does not impose a local neighborhood structure along the series index; locality is imposed only along the lag axis.

Importantly, none of the kernels performs a \emph{local} convolution along the series dimension; when the height is $N$, the filter spans all variables, and when the height is $1$, the filter is per-variable. The only notion of locality is along the lag (temporal) axis. This structured design enables the network to capture both global and local dependencies in multivariate time-series data. The convolutional feature extractors and subsequent fully connected layers are used in both the Focuser and Modeler branches to produce tensors with the desired transition representation shape. Concretely, for an input window $\bm{Q}_t \in \mathbb{R}^{N\times L}$, each branch produces a final linear output that is reshaped into a tensor in $\mathbb{R}^{N\times N\times L}$: this tensor is denoted $\bm{C}_t$ in the Modeler branch and $\bm{Z}_t$ in the Focuser branch. The first dimension indexes the target series, the second dimension indexes the source series, and the third dimension indexes the lag within the input window. This output structure is chosen so that the resulting tensor can be interpreted as a local transition tensor.

The concatenated representation $\bm{K}$ is then passed through a compact fully connected bottleneck. Let $d_1$ and $d_2$ denote the widths of the two hidden fully connected layers. In the reported experiments we use $d_1=128$ and $d_2=32$. For either branch, with branch-specific weights, the hidden representation is computed as
\begin{equation*}
\bm{h}_{1,t}
=
\tanh(\bm{W}_1\bm{K}+\bm{b}_1),
\qquad
\bm{h}_{1,t}\in\mathbb{R}^{d_1}.
\end{equation*}
\begin{equation*}
\bm{h}_{2,t}
=
\tanh(\bm{W}_2\bm{h}_{1,t}+\bm{b}_2),
\qquad
\bm{h}_{2,t}\in\mathbb{R}^{d_2}.
\end{equation*}
The branch output is then obtained by a final affine projection to $N^2L$ components, followed by reshaping to $N\times N\times L$:
\begin{equation*}
\bm{O}_t
=
\operatorname{reshape}_{N\times N\times L}
\!\left(
\bm{W}_{\mathrm{out}}\bm{h}_{2,t}+\bm{b}_{\mathrm{out}}
\right),
\qquad
\bm{O}_t\in\mathbb{R}^{N\times N\times L}.
\end{equation*}
Here $\bm{O}_t$ denotes the output tensor of one branch. In the Modeler branch, $\bm{O}_t=\bm{C}_t$; in the Focuser branch, $\bm{O}_t=\bm{Z}_t$, followed by $\bm{F}_t=\sigma_T(\bm{Z}_t)$. The widths $d_1=128$ and $d_2=32$ were used as a fixed compact bottleneck across all reported experiments. They are architectural hyperparameters and can be increased for larger systems or longer windows without changing the definition of the transition tensor or the interpretation procedure.

The tensor $\bm{C}_t$ is designed to encode the signed amplitude of the transition coefficients: it may take negative values, which correspond to inhibitory or anti-correlated effects, while larger magnitudes $|C_{n,i,\ell}|$ indicate stronger influence of source series $i$ at lag $\ell$ on the prediction of target series $n$. In the full transition tensor $\bm{\alpha}_t$, these signed amplitudes are combined with the corresponding Focuser mask, yielding per-sample, lag-resolved interaction patterns alongside the forecast.

The default architecture was selected after subset-selection checks over the convolutional branches and fully connected depth. We repeated each candidate configuration five times with different random initializations and assessed both forecasting performance and stability of the recovered coefficients. The final reported configuration uses all convolutional branches compatible with the chosen window length, 16 channels per active branch, and hidden widths $128$ and $32$.

For the Focuser, the branch output before the final activation is denoted $\bm{Z}_t$, i.e., the Focuser pre-activation tensor listed in Table~\ref{tab:shape_summary}. We apply a pointwise sigmoid function to this pre-activation tensor to produce the Focuser mask $\bm{F}_t$:
\begin{equation}\label{eq:focuser_sigmoid}
    \bm{F}_t=\sigma_T(\bm{Z}_t).
\end{equation}
Here, $\sigma_T(\cdot)$ denotes a temperature-dependent sigmoid,
\begin{equation}
\sigma_T(x)=\frac{1}{1+\exp(-x/T)}.
\end{equation}
The elements of $\bm{F}_t$ satisfy $F_{n,i,\ell}\in(0,1)$. The purpose of $\bm{F}_t$ is to suppress irrelevant source--lag entries and retain those that are useful for the current forecast. We set $T=1$ throughout.

The Modeler and Focuser differ primarily in their inputs: while the Focuser operates on the raw window $\bm{Q}_{t}$, the Modeler receives the masked input $\bm{F}_t \diamondsuit \bm{Q}_{t}$. Finally, the forecast is computed by multiplying the input window by both the Focuser mask and the Modeler coefficients:

Equivalently, using the explicit summation notation introduced in Eq.~\eqref{eq:transition_diamond}, the DCIts prediction is
\begin{equation}\label{eq:transition_nn1}
\hat X_{n,t+1}
=
\sum_{i=1}^{N}\sum_{\ell=1}^{L}
(C_t)_{n,i,\ell}\,(F_t)_{n,i,\ell}\,(Q_t)_{i,\ell}.
\end{equation}
Here, $\circ$ denotes the Hadamard product between tensors of the same shape,
\begin{equation}
(\bm{C}_t \circ \bm{F}_t)_{n,i,\ell}
=
(C_t)_{n,i,\ell}(F_t)_{n,i,\ell}.
\end{equation}
Comparing Eq.~\eqref{eq:transition_diamond} with Eq.~\eqref{eq:transition_nn1} gives
\begin{equation}
    \bm{\alpha}_t=\bm{C}_t\circ\bm{F}_t .
\end{equation}
The transition tensor is therefore the Hadamard product of the Focuser output $\bm{F}_t$ and the Modeler output $\bm{C}_t$. This construction differs from standard attention in both role and interpretation. The Focuser $\bm{F}_t$ acts as a sparse relevance mask, but the final explanatory object is the transition tensor $\bm{\alpha}_t=\bm{C}_t\circ\bm{F}_t$, whose entries are signed coefficients in the prediction equation rather than normalized attention scores. Thus, interpretability is tied directly to the predictive mechanism: the same coefficient tensor that is used to compute the forecast is also the object inspected to identify source series, lags, signs, and relative magnitudes of local interactions. Furthermore, tensors $\bm{F}_t$ and $\bm{C}_t$ are produced for each example, both during training and prediction. Consequently, $\bm{\alpha}_t$ is also sample-specific, allowing forecast-local inspection rather than only global, dataset-averaged interpretation. This preserves regime-dependent structure that would otherwise be washed out by aggregation.

\begin{figure}[htbp]
\centering
\scalebox{0.8}{%
\begin{minipage}{\textwidth}
\centering

\subfloat[]{%
\begin{minipage}[t]{0.45\textwidth}
\centering
\resizebox{\linewidth}{!}{%
\begin{tikzpicture}[node distance=0.6cm]

\node[box, fill=blue!10] (Input) at (0,0) {\textsc{Input}, $\bm{Q}_{t}$};
\node[box, fill=green!10, above=of Input] (Focuser) {\textsc{Focuser}, $\bm{F}_t=\sigma_T(\bm{Z}_t)$};
\node[box, fill=yellow!10, above=of Focuser] (X1) {$\bm{F}_t\diamondsuit\bm{Q}_{t}$};
\node[box, fill=red!10, above=of X1] (LinearModeler) {\textsc{Modeler}, $\bm{C}_t$};
\node[box, fill=yellow!10, above=of LinearModeler] (X2) {$(\bm{C}_t\circ\bm{F}_t)\diamondsuit\bm{Q}_{t}$};
\node[box, fill=blue!10, above=of X2] (Output) {\textsc{Output}};

\draw[arrow] (Input) -- (Focuser);
\draw[arrow] (Focuser) -- (X1);
\draw[arrow] (X1) -- (LinearModeler);
\draw[arrow] (LinearModeler) -- (X2);
\draw[arrow] (X2) -- (Output);

\draw[arrow, rounded corners] (Input) -| ([xshift=-0.2cm]Focuser.west) |- (X1.west);
\draw[arrow, rounded corners] (Input) -| ([xshift=-1.0cm]LinearModeler.west) |- (X2.west);
\draw[arrow, rounded corners] (Focuser) -| ([xshift=0.9cm]LinearModeler.east) |- (X2.east);

\end{tikzpicture}
}%
\end{minipage}%
}\hfill
\subfloat[]{%
\begin{minipage}[t]{0.5\textwidth}
\centering
\resizebox{\linewidth}{!}{%
\begin{tikzpicture}[node distance=0.45cm and 0.8cm]

\node[box, fill=blue!10, at={(0,0)}] (input) {\textsc{Input}};
\node[conv, above left=of input] (conv1) {\textsc{Conv}${}_1$};
\node[conv, above=of input] (conv2) {\ldots};
\node[conv, above right=of input] (conv3) {\textsc{Conv}${}_N$};
\node[box, fill=yellow!10, above=of conv2] (concat) {\textsc{Concat}};
\node[box, fill=gray!10, above=of concat] (linear1) {\textsc{Linear}${}_1$};
\node[op, above=of linear1] (tanh1) {\textsc{Tanh}};
\node[draw=none, fill=none, above=of tanh1] (linear2) {\ldots};
\node[box, fill=gray!10, above=of linear2] (linear3) {\textsc{Linear}${}_N$};
\node[op, above=of linear3] (tanh3) {\textsc{Tanh}};
\node[box, fill=red!20, above=of tanh3] (linear4) {\textsc{Linear output} $\bm{C}_t$ or $\bm{Z}_t$};

\begin{scope}[on background layer]
  \node[draw, rounded corners=10pt, line width=1.5pt, inner sep=10pt,
        fit={(-4,-0.5) (4,12.5)}] (outerbox) {};
  \node[draw,  rounded corners=10pt, line width=1.5pt, inner sep=10pt,
        fit={(input) (conv1) (conv2) (conv3) (concat) (linear1) (tanh1) (linear2) (linear3) (tanh3) (linear4)}] (innerbox) {};
\end{scope}

\draw[arrow, rounded corners] (input.west) -| (conv1.south);
\draw[arrow] (input) -- (conv2.south);
\draw[arrow, rounded corners] (input.east) -| (conv3.south);
\draw[arrow, rounded corners] (conv1.north) |- (concat.west);
\draw[arrow] (conv2.north) -- (concat.south);
\draw[arrow, rounded corners] (conv3.north) |- (concat.east);
\draw[arrow] (concat) -- (linear1);
\draw[arrow] (linear1) -- (tanh1);
\draw[arrow] (tanh1) -- (linear2);
\draw[arrow] (linear2) -- (linear3);
\draw[arrow] (linear3) -- (tanh3);
\draw[arrow] (tanh3) -- (linear4);

\node[below right=2pt and 2pt, font=\Large] at (outerbox.north west)
{\textsc{Modeler/Focuser branch}};

\end{tikzpicture}
}%
\end{minipage}%
}

\end{minipage}%
}%
\caption{Overview of the DCIts architecture after the quantities used in the schematic have been defined. 
(a) Two-step prediction mechanism. The input window $\bm{Q}_t$ is first passed through the Focuser branch to obtain a pre-activation tensor $\bm{Z}_t$, followed by the mask $\bm{F}_t=\sigma_T(\bm{Z}_t)$, where $\sigma_T$ is the temperature-dependent sigmoid. The masked input $\bm{F}_t\diamondsuit\bm{Q}_t$ is then passed through the Modeler to obtain the signed coefficient tensor $\bm{C}_t$. The final transition tensor is $\bm{\alpha}_t=\bm{C}_t\circ\bm{F}_t$, and the forecast is obtained from $(\bm{C}_t\circ\bm{F}_t)\diamondsuit\bm{Q}_t$. Here, $\circ$ denotes the Hadamard product between tensors of the same shape, and $\diamondsuit$ denotes the broadcasted element-wise product with the input window. 
(b) Shared convolutional and fully connected branch architecture. The same architectural template is used in both branches: in the Modeler branch, the final linear output is interpreted as the signed coefficient tensor $\bm{C}_t$; in the Focuser branch, the corresponding final linear output is interpreted as the pre-activation tensor $\bm{Z}_t$, and the mask is obtained as $\bm{F}_t=\sigma_T(\bm{Z}_t)$.}
\label{ARH}
\end{figure}

\subsection{Interpretability}

We refer to the transition tensor $\bm{\alpha}_t$ to maintain consistency with the notation used in related literature. We will also refer to the transition tensor as $\bm{\alpha}_t$ coefficients when discussing model interpretability. Because the model computes a transition tensor for each windowed sample, a dataset of $M-L+1$ samples yields $M-L+1$ corresponding transition tensors $\bm{\alpha}_t$, indexed by the sample.
 Throughout the remainder of this paper, unless otherwise specified, any reference to $\bm{\alpha}_t$ coefficients or derived quantities pertains to the coefficients from the individual sample $\bm{\alpha}_t$, $\bm{\alpha}_t \in \mathbb{R}^{N \times N \times L}$. We emphasize that the model does not learn a separate, unconstrained parameter $\bm{\alpha}_t$ for each time step, which would lead to an exponential number of free parameters and trivial data fitting. DCIts is parameterized as a single neural network; $\bm{\alpha}_t$ is the output of this network given input $\bm{Q}_t$, rather than a free parameter.

The model's $\bm{\alpha}_t$ coefficients quantify the influence of all time series, including self-influence, on the final output at each time point within the window, i.e., modeling the next step in the time series, as discussed previously. In practice, interpretability extraction follows directly from the trained network: for each test window $\bm{Q}_{t}$ we run a forward pass, obtain the Focuser and Modeler tensors $(\bm{F}_t,\bm{C}_t)$, and form the transition tensor $\bm{\alpha}_t=\bm{F}_t\circ\bm{C}_t$. Our goal is therefore a \emph{predictive competitiveness check} under a controlled setting (same generators, same horizon, and matched window length), rather than an exhaustive leaderboard across forecasting families.

To focus solely on the mutual influence among time series, while setting aside the temporal aspect of their interactions, we introduce the $\tilde{\bm{\beta}}$ and $\bm{\beta}$ coefficients. Both $\tilde{\bm{\beta}}$ and $\bm{\beta}$ have dimensions $\mathbb{R}^{N \times N}$, allowing us to analyze the relative impact between time series without directly considering the time lag of these interactions:

\begin{equation}
    \tilde{\beta}_{i,j} = \sum_{\ell=1}^{L} |\alpha_{i,j,\ell}|,
\end{equation}
\begin{equation}
    \beta_{i,j} = \frac{\tilde{\beta}_{i,j}}{\sum_{k=1}^{N} \tilde{\beta}_{i,k}}.
\end{equation}

The key difference between $\tilde{\bm{\beta}}$ and $\bm{\beta}$ lies in the normalization applied to the latter. Specifically, $\bm{\beta}$ is normalized per time series, meaning that $\beta_{i,j}$ represents the proportional impact of time series $j$ on time series $i$. This effectively quantifies the relative influence among time series. In contrast, $\tilde{\bm{\beta}}$ is simply the aggregate of the $\bm{\alpha}$ coefficients across all lags and does not provide insight into how the influence is distributed among the different time series.

Using the $\bm{\beta}$ coefficients, we can determine which time series are the key drivers behind each series in the system. Although the absolute value in the equation loses the interaction sign, the magnitude of each time series's impact on the other remains clear. Moreover, the $\bm{\beta}$ values are intuitive and readily understandable to a broad audience, not just domain experts or machine learning practitioners. Higher $\bm{\beta}$ values indicate stronger influence, while lower $\bm{\beta}$ values suggest weaker or negligible influence.

In addition, if the goal is to understand how the data is generated, i.e., to uncover the underlying equations governing the MTS system, we can refer back to the $\bm{\alpha}_t$ coefficients. These coefficients provide a direct way to reconstruct the generating equations. Ultimately, DCIts is used not only to predict the next step in the time series but also to infer the equations that drive the data. By analyzing both the $\bm{\alpha}$ and $\bm{\beta}$ coefficients, we gain a deeper understanding of the system's dynamics and structure.

We opted to train the model on a forecasting task, under the conviction that this approach, as opposed to direct optimization for interpretability, fosters more genuine and predictive, lag-resolved insights interpretable in terms of influence. Typically, models and methods that impose excessive structure on anticipated interpretability can distort or entirely overlook it. Our approach minimally requires linear interpretability and then adds higher-order terms as necessary, aligning with conventional modeling practices in physics. Several considerations drive this choice. We contend that designing the model architecture with the forecasting task in mind not only enhances the model's predictive capability but also ensures that the approximation closely mirrors the underlying equations generating the data. This alignment is crucial, as it allows the model to generate interpretability concurrently with forecasting. This means the model is equipped to determine whether different equations govern the data for specific input values. Our aim in the present work is more specific: to test whether a neural forecasting model can expose a sample-specific transition tensor whose coefficients remain faithful to known source--lag interactions in controlled multivariate benchmarks.

\subsection{Run-to-run stability filtering of coefficients}
\label{sec:stability_filter}

To assess the robustness of the learned lag-resolved coefficients to random initialization, we repeat training $R$ times with identical hyperparameters and data splits but different random seeds. Let $(\alpha_{(r)})_{n,i,\ell}$ denote any reported coefficient entry (target $n$, source $i$, lag $\ell$) obtained from run $r\in\{1,\dots,R\}$. We compute the across-run mean and standard deviation,
\begin{align}
\mu_{n,i,\ell} &= \frac{1}{R}\sum_{r=1}^{R}(\alpha_{(r)})_{n,i,\ell},\\
\sigma_{n,i,\ell} &= \sqrt{\frac{1}{R-1}\sum_{r=1}^{R}\bigl((\alpha_{(r)})_{n,i,\ell}-\mu_{n,i,\ell}\bigr)^2}.
\end{align}
For visualization and qualitative comparison to ground truth on synthetic systems, we retain only ``stable'' entries whose magnitude consistently exceeds the run-to-run variability:
\begin{equation}
\mathbb{I}_{n,i,\ell}
=
\mathbf{1}\!\left(
|\mu_{n,i,\ell}| > c\,\sigma_{n,i,\ell}
\right),
\label{eq:stability_mask}
\end{equation}
with $c=1.95$. 
The factor $c$ suppresses coefficients that are not reproducible across random initializations. We use this criterion as a conservative robustness filter for interpretability plots, not as a formal hypothesis test across the full tensor.

\subsection{Optimal Window Size}

In DCIts, the window size $L$ is most naturally interpreted as an \emph{upper bound} on the memory depth rather than a fragile hyperparameter that must be finely tuned. This is possible because the model outputs a lag-resolved transition tensor $\bm{\alpha}_t$ whose entries can be assessed for \emph{run-to-run robustness} under different random initializations.

We fix a conservatively large $L_{\max}$ and train DCIts $R$ times with identical data splits and hyperparameters but different random seeds. We compute the across-run mean and standard deviation, and retain only entries that are stable across random initializations (cf. Sec.~\ref{sec:stability_filter}). This criterion suppresses coefficients that are not reproducible across runs and, therefore, are unlikely to reflect a robust lag-dependent interaction.

Having fixed $L=L_{\max}$, we then perform a \emph{step-by-step lag inspection}: we examine the masks $\mathbb{I}_{:,:,1},\mathbb{I}_{:,:,2},\dots,\mathbb{I}_{:,:,L_{\max}}$ over (target, source) pairs and identify the largest lag $\ell^\star$ for which a non-negligible set of entries remains stable. In this way, $\ell^\star$ provides an empirical estimate of the system's effective memory depth, obtained directly from the model-internal representation, without requiring an explicit hyperparameter search over $L$. In practice, lags $\ell>\ell^\star$ are automatically attenuated by instability (large $\sigma$ relative to $|\mu|$) .

For completeness and to align with standard forecasting practice, one may also plot the average predictive loss as a function of $L$. Specifically, for each candidate $L\in\{1,\dots,L_{\max}\}$ we train the model $R$ times and compute the mean test performance $\overline{\mathcal{L}}(L)$ (e.g., MSE). This curve typically exhibits a pronounced reduction once $L$ is sufficiently large to encompass the dominant lag structure, followed by saturation for larger $L$. We include such a plot later solely as a diagnostic illustration; the recommended procedure in DCIts is to set $L_{\max}$ conservatively and infer the effective lag support from the run-to-run stability of $\bm{\alpha}_t$.

\subsection{Higher order approximations}\label{higher-order}

The coefficient construction above corresponds to the lag-resolved linear transition term ($p=1$). We now describe the optional extension that adds a per-target bias term ($p=0$) and elementwise higher-order polynomial terms ($p\geq2$). The architecture of this extended DCIts formulation is shown in Fig.~\ref{DCIts_nonlinear_arh}.

In this section, we denote the input window $\bm{Q}_{t}$ by $\bm{q}=\bm{Q}_{t}\in\mathbb{R}^{N\times L}$ for readability, where $N$ is the number of series and $L$ is the window length (lag depth).

\paragraph{Bias input (order $p=0$).}
To represent a per-target additive bias, we introduce a constant bias input
\begin{equation}
\bm{q}^{(0)} \in \mathbb{R}^{N\times 1},
\qquad
(\bm{q}^{(0)})_{i,1}=1,\ \ i=1,\dots,N,
\end{equation}
i.e., a column of ones with lag depth $L_0=1$.
The corresponding bias coefficients are encoded in a tensor
$\bm{\alpha}^{(0)}_t\in\mathbb{R}^{N\times N\times 1}$ that is constrained to be diagonal in the (target, source) indices,
\begin{equation}
\alpha^{(0)}_{n,i,1}=b_{n,t}\,\delta_{ni},
\end{equation}
where $\delta_{ni}$ is the Kronecker delta, i.e., $\delta_{ni}=1$ if $n=i$ and $\delta_{ni}=0$ otherwise. This diagonal constraint gives a minimal implementation of the bias term: the order-$p=0$ contribution reduces to one additive bias $b_{n,t}$ for each target series $n$. More general bias tensors could be introduced, but we do not use them here in order to keep the bias branch compact and directly interpretable.
Here, $b_{n,t}$ is sample-dependent (produced by the network from the current input window), not a single global scalar.

\paragraph{Elementwise polynomial terms (orders $p\ge 1$).}
For higher-order contributions, we use elementwise powers of the window:
\begin{equation}
\bm{q}^{(p)} \in\mathbb{R}^{N\times L},
\qquad
(\bm{q}^{(p)})_{i,\ell} = (q_{i,\ell})^p,\ \ p\ge 1,
\end{equation}
so the order-$p$ term represents a monomial of a single source--lag entry, i.e., an elementwise power rather than a tensor or matrix power. Consequently, the present higher-order extension can represent signed, lag-resolved contributions of the form $\alpha^{(p)}_{n,i,\ell}\,(q_{i,\ell})^p,$
including linear terms ($p=1$) and elementwise polynomial terms ($p>1$). It does not, in its current form, include mixed products across distinct source--lag entries, such as $q_{i,\ell_1}q_{j,\ell_2}$ with $i\neq j$ and/or $\ell_1\neq\ell_2$. This is a deliberate design choice in the present version: excluding mixed cross-terms keeps the coefficient tensor compact, avoids a combinatorial expansion of interaction terms, and preserves direct inspection of target--source--lag contributions. Mixed multiplicative interactions could be incorporated in future extensions by adding explicit cross-term branches, but they are outside the scope of the present model.

For each order $p$ we define a lag-resolved transition tensor
$\bm{\alpha}^{(p)}_t\in\mathbb{R}^{N\times N\times L_p}$, with $L_0=1$ and $L_p=L$ for $p\ge 1$.
As in the linear model, we factorize
\begin{equation}
\bm{\alpha}^{(p)}_t = \bm{C}^{(p)}_t \circ \bm{F}^{(p)}_t,
\end{equation}
where $\bm{C}^{(p)}_t$ provides signed coefficients and $\bm{F}^{(p)}_t$ is the corresponding Focuser mask. Analogously to the linear case, the Focuser branch first produces a pre-activation tensor $\bm{Z}^{(p)}_t$, and the mask is obtained as $\bm{F}^{(p)}_t=\sigma_T(\bm{Z}^{(p)}_t)$.

The extended one-step transition is then
\begin{equation}\label{eq:transition_with_bias}
X_{n,t+1}
=
b_{n,t}
\;+\;
\sum_{p=1}^{P}\sum_{i=1}^{N}\sum_{\ell=1}^{L}
\bigl(\bm{\alpha}^{(p)}_t \diamondsuit \bm{q}^{(p)}\bigr)_{n,i,\ell}.
\end{equation}
Here, $\diamondsuit$ denotes the broadcasted element-wise (Hadamard) product:
\begin{equation}
\bigl(\bm{\alpha}^{(p)}_t \diamondsuit \bm{q}^{(p)}\bigr)_{n,i,\ell}
=
\alpha^{(p)}_{n,i,\ell}\, (q^{(p)})_{i,\ell},
\end{equation}
with $\ell=1,\dots,L$ for $p\ge 1$.
This formulation allows DCIts to capture a per-target bias and interpretable, lag-resolved elementwise polynomial effects in a unified framework. It should not be read as a complete multivariate polynomial expansion, because mixed cross-products between different source--lag entries are not included in the present implementation.

\begin{figure}[htb]
\centering
\resizebox{\textwidth}{!}{
\begin{tikzpicture}
\tikzset{
  base/.style={draw, rounded corners, line width=0.8pt, inner sep=4pt},
  rect2x1/.style={base, minimum width=2.2cm, minimum height=1cm},
  rect1x1/.style={base, minimum width=1.2cm, minimum height=1cm},
  circ/.style={base, circle, minimum size=1cm},
  arrow/.style={->, line width=0.8pt, >={Latex[round,open]}}
}

\node[rect2x1, fill=blue!10] (Input) at (0,0) {\textsc{Input}, $\bm{q}$};

\node[rect2x1, fill=yellow!10] (qin1) at (3.8,-4) {$\bm{q}^{(p)}$};
\node[rect2x1, fill=yellow!10] (qin2) at (3.8,-2) {$\vdots$};
\node[rect2x1, fill=yellow!10] (qin3) at (3.8,0) {$\bm{q}$};
\node[rect2x1, fill=yellow!10] (qin4) at (3.8,2) {$\bm{q}^{(0)}=\mathbf{1}$};

\node[rect1x1, fill=green!10] (Fin1) at (7.0,-4) {$\bm{F}_p$};
\node[rect1x1, fill=green!10] (Fin2) at (7.0,-2) {$\vdots$};
\node[rect1x1, fill=green!10] (Fin3) at (7.0,0) {$\bm{F}_1$};
\node[rect1x1, fill=green!10] (Fin4) at (7.0,2) {$\bm{F}_0$};

\node[rect2x1, fill=yellow!10] (Xin1) at (10.8,-4) {$\bm{F}_p\diamondsuit \bm{q}^{(p)}$};
\node[rect2x1, fill=yellow!10] (Xin2) at (10.8,-2) {$\vdots$};
\node[rect2x1, fill=yellow!10] (Xin3) at (10.8,0) {$\bm{F}_1\diamondsuit \bm{q}$};
\node[rect2x1, fill=yellow!10] (Xin4) at (10.8,2) {$\bm{F}_0\diamondsuit \bm{q}^{(0)}$};

\node[rect1x1, fill=red!10] (Model1) at (14.2,-4) {$\bm{C}_p$};
\node[rect1x1, fill=red!10] (Model2) at (14.2,-2) {$\vdots$};
\node[rect1x1, fill=red!10] (Model3) at (14.2,0) {$\bm{C}_1$};
\node[rect1x1, fill=red!10] (Model4) at (14.2,2) {$\bm{C}_0$};

\node[rect2x1, fill=yellow!10] (X1) at (17.8,-4) {$(\bm{C}_p \circ \bm{F}_p)\diamondsuit \bm{q}^{(p)}$};
\node[rect2x1, fill=yellow!10] (X2) at (17.8,-2) {$\vdots$};
\node[rect2x1, fill=yellow!10] (X3) at (17.8,0) {$(\bm{C}_1 \circ \bm{F}_1)\diamondsuit \bm{q}$};
\node[rect2x1, fill=yellow!10] (X4) at (17.8,2) {$(\bm{C}_0 \circ \bm{F}_0)\diamondsuit \bm{q}^{(0)}$};

\node[circ, fill=yellow!10] (Plus) at (21.4,0) {+};
\node[rect2x1, fill=blue!10] (Output) at (24.0,0) {\textsc{Output}};

\node[draw, dashed, inner sep=12pt, fit=(qin1) (qin4)] (BoxIn1) {};
\node[draw, dashed, inner sep=12pt, fit=(Xin1) (Xin4)] (BoxIn2) {};
\node[draw, dashed, inner sep=12pt, fit=(Fin1) (Fin4)] (BoxF) {};
\node[draw, dashed, inner sep=12pt, fit=(X1) (X4)] (BoxOut) {};

\draw[arrow, rounded corners] ([xshift=0.4cm]Input.east) |- (qin1.west);
\draw[arrow] (Input.east) -- (qin3.west);

\draw[arrow] (qin1.east) -- (Fin1.west);
\draw[arrow] (qin3.east) -- (Fin3.west);
\draw[arrow] (qin4.east) -- (Fin4.west);

\draw[arrow] (Fin1.east) -- (Xin1.west);
\draw[arrow] (Fin3.east) -- (Xin3.west);
\draw[arrow] (Fin4.east) -- (Xin4.west);

\draw[arrow] (Xin1.east) -- (Model1.west);
\draw[arrow] (Xin3.east) -- (Model3.west);
\draw[arrow] (Xin4.east) -- (Model4.west);

\draw[arrow] (Model1.east) -- (X1.west);
\draw[arrow] (Model3.east) -- (X3.west);
\draw[arrow] (Model4.east) -- (X4.west);

\draw[arrow, rounded corners] (X1.east) -| (Plus.west);
\draw[arrow] (X3.east) -- (Plus.west);
\draw[arrow, rounded corners] (X4.east) -| (Plus.west);

\draw[arrow] (Plus.east) -- (Output.west);

\end{tikzpicture}
}
\caption{Extended DCIts architecture with bias ($p=0$) and elementwise polynomial terms ($p\ge 1$). The bias branch uses a constant input $\bm{q}^{(0)}=\mathbf{1}$ with lag depth $L_0=1$ and yields a per-target additive bias via diagonal coefficients. Higher-order branches use elementwise powers $\bm{q}^{(p)}$ and produce lag-resolved contributions through $\bm{\alpha}^{(p)}_t=\bm{C}^{(p)}_t\circ \bm{F}^{(p)}_t$.}
\label{DCIts_nonlinear_arh}
\end{figure}

The notation for the $\bm{\alpha}_t$, $\tilde{\bm{\beta}}$, and $\bm{\beta}$ coefficients is adjusted to
$\bm{\alpha}^{(p)}_t$, $\tilde{\bm{\beta}}^{(p)}$, and $\bm{\beta}^{(p)}$ to reflect interpretability for a specific order $p$.
Unless otherwise stated, writing $\bm{\alpha}_t$, $\tilde{\bm{\beta}}$, and $\bm{\beta}$ implies $p=1$.

\section{Assessing Model Performance and Interpretability Accuracy}\label{sec-experiments}

This section examines the interpretability, accuracy, and forecasting performance of the DCIts architecture. Initially, the focus is on forecasting performance; then, the impact of different generating processes for test data on model performance and interpretability is examined. The general code is written in Python, and the DCIts neural network is implemented using PyTorch. The complete code, along with examples, is accessible on GitHub \cite{githubdcits}. It is important to note that model training and inference on benchmark datasets are not hardware-intensive.

\subsection{Reproducibility protocol: datasets, split, windowing, seeds, and stopping}
\label{sec:repro_protocol}

\paragraph{Contiguous split before windowing (no leakage).}
Let $\bm{X}\in\mathbb{R}^{N\times M}$ denote the full multivariate series of length $M$.
We split $\bm{X}$ \emph{along time} into three contiguous, non-overlapping segments
$\bm{X}_{\mathrm{tr}}, \bm{X}_{\mathrm{val}}, \bm{X}_{\mathrm{te}}$ using ratios $(0.6,0.2,0.2)$ by default.
To ensure that each split yields at least one supervised example for window length $L$, we enforce
$\lvert \bm{X}_{\mathrm{tr}}\rvert,\lvert \bm{X}_{\mathrm{val}}\rvert,\lvert \bm{X}_{\mathrm{te}}\rvert \ge L+1$.
If a ratio-based split violates this constraint, the affected split(s) are increased to length $L+1$ and the remaining length is deducted from the other split(s), while preserving contiguity and
$\lvert \bm{X}_{\mathrm{tr}}\rvert+\lvert \bm{X}_{\mathrm{val}}\rvert+\lvert \bm{X}_{\mathrm{te}}\rvert=M$.

\paragraph{Window construction and shuffling.}
After splitting, we construct rolling windows \emph{within each segment only}.
From a segment $\bm{X}_\star$ of length $m$, we form inputs $\bm{Q}_t\in\mathbb{R}^{N\times L}$ for
$t=L,\dots,m-1$, with targets $\bm{X}_{\star,:,t+1}$.
We shuffle only the set of training windows $\bm{Q}_{\mathrm{tr}}$; validation and test windows preserve temporal order.

\paragraph{Repeated runs and random seeds.}
All reported results are aggregated across $R$ independent training runs with identical data splits and hyperparameters, but different random seeds (which affect parameter initialization and mini-batch ordering).
Synthetic data generation uses a fixed seed so that all compared methods observe identical realizations.

\paragraph{Optimization and early stopping.}
Unless stated otherwise, models are trained with Adam using learning rate $10^{-3}$, batch size 64, and mean squared error (MSE) as the default forecasting loss. In selected analyses, we repeat training with mean absolute error (MAE/L1 loss) to check the sensitivity of coefficient recovery to the loss function. Training runs for at most $E_{\max}=100$ epochs. The learning-rate scheduler uses patience $p_{\mathrm{sched}}=5$. Early stopping is controlled by an early-stopping modifier equal to 2, corresponding to two scheduler-patience intervals without validation improvement, i.e., 10 epochs without improvement in the default setting. The model parameters with the best validation loss are restored. Here, ``improve'' means a decrease relative to the best validation loss observed so far.

\paragraph{Datasets and baseline.}
The benchmarking datasets were sourced from Bari\'c et al.~\cite{baric2021}, with minor typographical errors corrected to match the generating code as described in that reference.
In addition, the generating process for Dataset~8 was adjusted so that the first time series contains the switching process.
Full dataset specifications are given in Appendix~\ref{bdata}.
\paragraph{Model variants used in the experiments.}
Unless explicitly stated otherwise, the benchmark comparison with IMV-LSTM uses the default bias-plus-linear DCIts configuration, corresponding to active orders $p=0$ and $p=1$ only. The bias branch is included so that the model can determine whether a per-target additive term is needed; bias terms are reported only when they are stable and non-negligible under the repeated-run interpretation procedure. Otherwise, the reported interpretability quantities focus on the $p=1$ lag-resolved transition coefficients. The training routine stores the test-set Focuser and Modeler tensors, which are then combined as $\bm{\alpha}_t=\bm{C}_t\circ\bm{F}_t$ for interpretation. Higher-order DCIts refers to configurations with polynomial branches $p\geq2$ activated; these are used only in experiments designed to test higher-order recovery.

To assess whether intrinsic coefficient-level interpretability comes at a substantial cost in forecasting accuracy, we use IMV-LSTM~\cite{guo2019exploring} as the main deep interpretable baseline. This choice follows a prior baseline-selection step. In our previous benchmark on the same family of synthetic generators, Bari\'c et al.~\cite{baric2021} evaluated several attention-based or interpretable models, including Seq2Graph, IMV-LSTM, TCDF, and DA-RNN, and found IMV-LSTM to be the strongest relevant baseline when both forecasting performance and interpretability reliability are considered. We therefore use IMV-LSTM as the primary reference model for testing whether DCIts can provide intrinsic coefficient-level interpretability without sacrificing predictive competitiveness. Our goal is a controlled predictive-competitiveness check under matched generators, horizon, and window length, rather than an exhaustive forecasting leaderboard across all modern architectures. Nevertheless, to verify that the robustness conclusions are not an artifact of comparison with a single baseline, we also performed additional forecasting-only robustness checks against the other attention-based baselines from the same benchmark family, including DA-RNN, TCDF, and Seq2Graph. In the Dataset~7 high-noise forecasting-only check, DCIts achieved lower MSE than IMV-LSTM, TCDF, and Seq2Graph, while DA-RNN achieved lower MSE in that particular forecasting-only test; these results were used only as robustness context and not as interpretability evidence. These additional checks are summarized only where directly relevant, because the central contribution of DCIts is not universal forecasting superiority but the combination of competitive prediction with an explicit, signed, lag-resolved, sample-specific coefficient tensor. Accordingly, the conclusions below are restricted to this controlled benchmark setting and to the stated interpretable-baseline comparisons.

\paragraph{Why synthetic benchmarks.}
When working with real-world multivariate time-series data, the data-generating mechanism is typically unknown, so the ``true'' directed, signed, and lag-resolved interaction structure cannot be observed directly. Consequently, the \emph{faithfulness} of learned explanations is hard to validate quantitatively and is often assessed only qualitatively or via indirect proxies. This motivates evaluation on synthetic generators with fully specified ground-truth coefficients and regimes, which is also standard in time-series causal/dependency benchmarking when controlled validation is required~\cite{Rungeetal19}. 

We therefore use the curated synthetic benchmark suite of Bari\'c et al.~\cite{baric2021}, which was designed to provide lag-resolved ground truth for testing interpretable multivariate time-series methods. The suite spans a systematic progression of dynamical complexity, including autoregressive, cross-coupled, nonlinear, stochastic, and switching/regime-dependent processes. Importantly, these benchmarks are not restricted to simple stationary deterministic examples: stochastic forcing is included, and some autoregressive cases have unit-sum lag coefficients, yielding random-walk-like or unit-root-like behavior under noise.

This controlled setting allows us to assess simultaneously forecasting performance and the ability of DCIts to recover the correct source series, lags, signs, and interaction orders, including the stability of coefficient recovery under controlled perturbations such as noise level and dimensionality.

While real-world case studies remain essential for demonstrating practical utility, synthetic benchmarks are the appropriate setting for \emph{quantitative} validation of coefficient-level interpretability claims and for method comparison on equal footing~\cite{fauvel2020performance,baric2021}. The present results should therefore be interpreted as controlled validation of the proposed interpretability mechanism, not as a complete demonstration of generalization to all real-world time-series settings.

\subsection{Quantitative analysis}\label{quant-analysis}
We examine the forecasting error, quantified by the Mean Squared Error (MSE), across the eight datasets utilized in Barić et al.~\cite{baric2021}. The comparison with IMV-LSTM uses the default bias-plus-linear DCIts configuration, i.e., active orders $p=0$ and $p=1$ only. The stability of errors across multiple iterations on the same dataset is analyzed, and the variation in error rates in response to changes in data-generation parameters is explored.

In Table~\ref{tab:mse_benchmark}, we report mean MSE (averaged over $R$ runs) for DCIts and IMV-LSTM. Across the eight generators, DCIts achieves a predictive error comparable to that of IMV-LSTM, indicating that the explicit coefficient construction does not incur a prohibitive accuracy cost. We emphasize that these results should be interpreted as establishing \emph{predictive competitiveness within this controlled benchmark suite}, not as a claim of universal superiority over all modern forecasting architectures. Unless stated otherwise, we set $L=10$ to match the configuration used for IMV-LSTM and to ensure a direct comparison.

\begin{figure}[htb]
\centering
\includegraphics[width=13 cm]{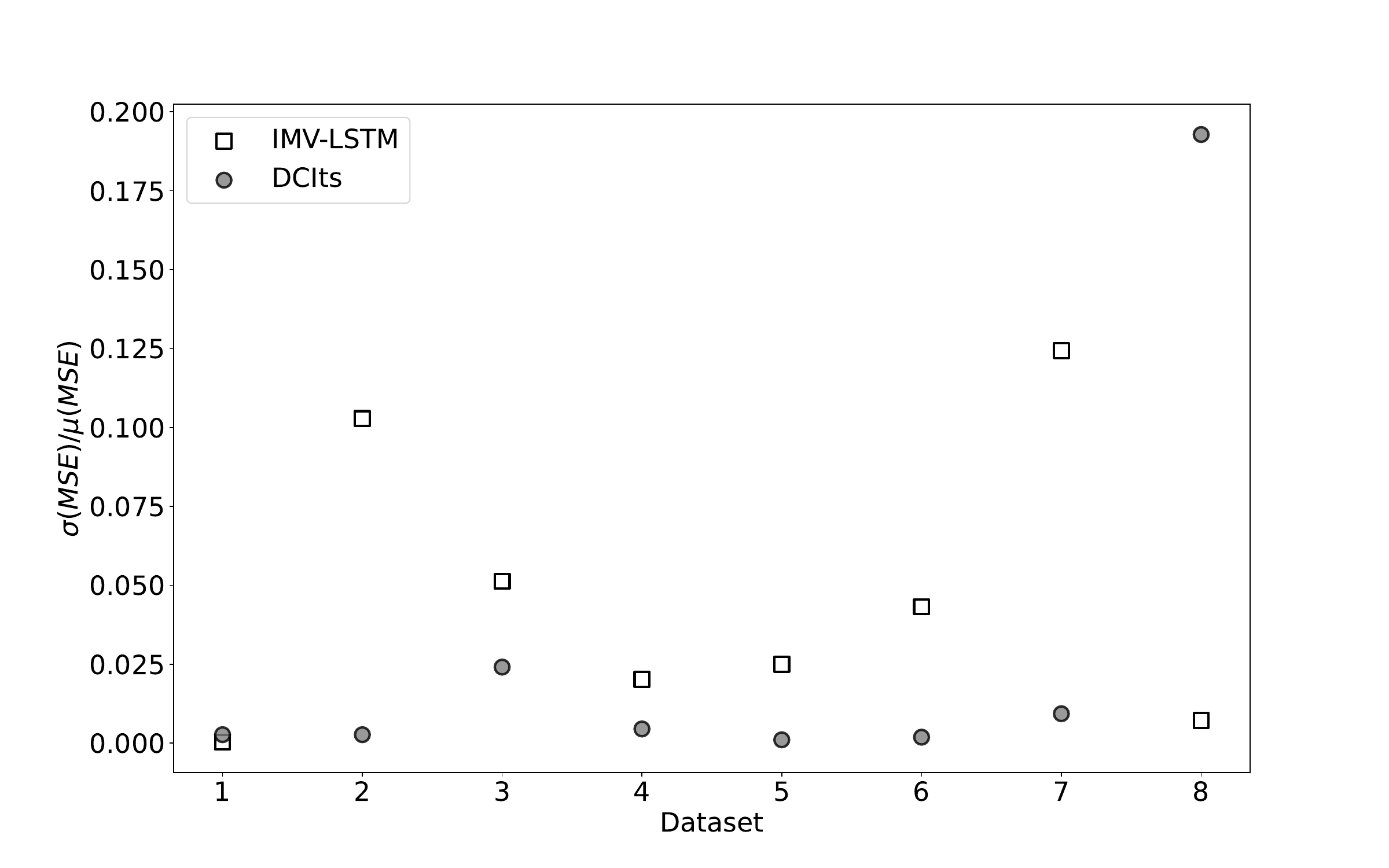}
\caption{The stability of prediction performance by dataset is assessed by plotting the standard deviation of the Mean Squared Error (MSE) divided by the mean MSE value on the y-axis. The x-axis represents the index of each dataset.}
\label{Performance stability}
\end{figure}

This distinction in performance is also reflected in the models' stability, analyzed in Figure \ref{Performance stability}. Here, the standard deviation of the experiment error, divided by the average experiment error, illustrates the stability of each model. Circles represent DCIts, and squares represent IMV-LSTM. Both models exhibit comparable performance, with slight differences in stability. Overall, DCIts shows comparable performance and stability to IMV-LSTM, and achieves lower error and/or lower relative variability on several datasets. In Dataset 8, IMV-LSTM appears to show greater stability; however, this is because the DCIts results are an order of magnitude smaller, making minor variations more prominent relative to them.

All Datasets 1--8 were generated following the processes used in~\cite{baric2021}, with the corrected specifications given in Appendix~\ref{bdata}. Unless otherwise stated, each synthetic realization has length $M=20000$ after discarding a burn-in of 1000 points. Initial values are sampled from a standard normal distribution, and values are then generated sequentially according to the corresponding data-generating equation. Additive Gaussian noise is injected intermittently: at each time step, a value $p\sim \mathrm{Uniform}(0,1)$ is sampled and
\begin{equation*}
\epsilon_t =
\begin{cases}
\mathcal{N}(0,\sigma_{\mathrm{noise}}^2), & p<f,\\
0, & \text{otherwise}.
\end{cases}
\end{equation*}
Unless otherwise stated, the noise frequency is $f=0.3$, the noise mean is zero, and the noise-variance parameter is $\sigma^2_{\mathrm{noise}}=0.1$. The stability of prediction performance is evaluated by the ratio of the standard deviation to the mean of the Mean Squared Error (MSE); lower values indicate greater stability.

\begin{table}[htpb]
\centering
\begin{tabular}{|c|c|c|}
\hline
Dataset & DCIts & IMV-LSTM \\
\hline
1 & \( (2.98 \pm 0.01) \times 10^{-3} \) & \( (2.91 \pm 0.01) \times 10^{-3} \) \\ \hline
2 & \( (7.58 \pm 0.02) \times 10^{-4} \) & \( (1.81 \pm 0.19) \times 10^{-3} \) \\ \hline
3 & \( (6.17 \pm 0.15) \times 10^{-5} \) & \( (1.25 \pm 0.06) \times 10^{-4} \) \\ \hline
4 & \( (1.23 \pm 0.01) \times 10^{-4} \) & \( (2.38 \pm 0.05) \times 10^{-4} \) \\ \hline
5 & \( (1.25 \pm 0.01) \times 10^{-3} \) & \( (1.38 \pm 0.03) \times 10^{-3} \) \\ \hline
6 & \( (1.27 \pm 0.02) \times 10^{-3} \) & \( (1.43 \pm 0.06) \times 10^{-3} \) \\ \hline
7 & \( (3.10 \pm 0.03) \times 10^{-3} \) & \( (5.14 \pm 0.64) \times 10^{-3} \) \\ \hline
8 & \( (2.55 \pm 0.49) \times 10^{-2} \) & \( (2.58 \pm 0.02) \times 10^{-1} \) \\ \hline
\end{tabular}
\caption{Benchmark forecasting performance (MSE) on Datasets 1--8~\cite{baric2021}. Reported values are the mean $\pm$ standard deviation over $R$ runs with different random seeds. The DCIts column corresponds to the default bias-plus-linear configuration, with active orders $p=0$ and $p=1$. DCIts exhibits accuracy comparable to IMV-LSTM on this controlled suite, while providing intrinsic, lag-resolved local interpretability.}
\label{tab:mse_benchmark}
\end{table}

\begin{figure}[htb]
\centering
\includegraphics[width=6.7 cm]{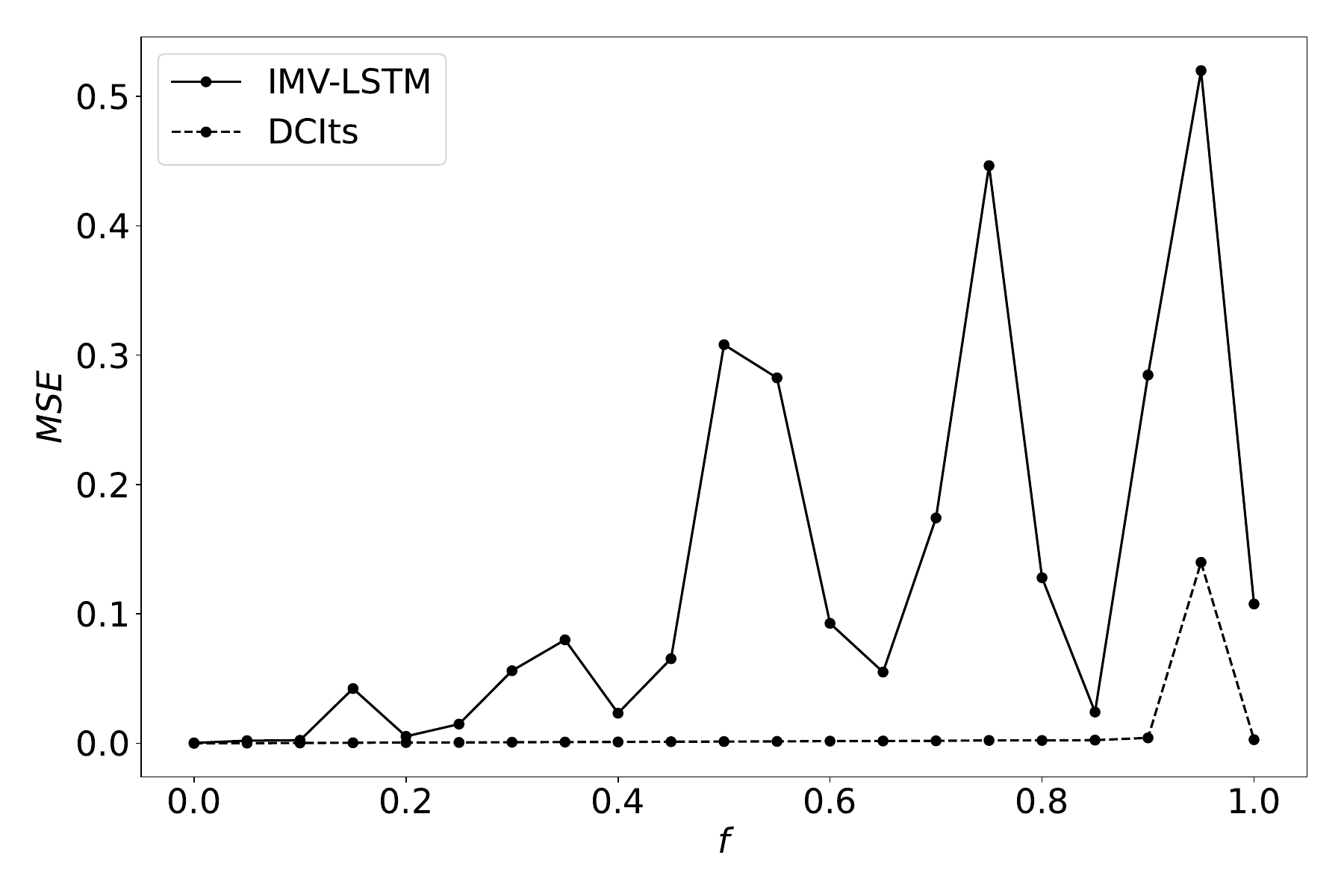}
\includegraphics[width=6.7 cm]{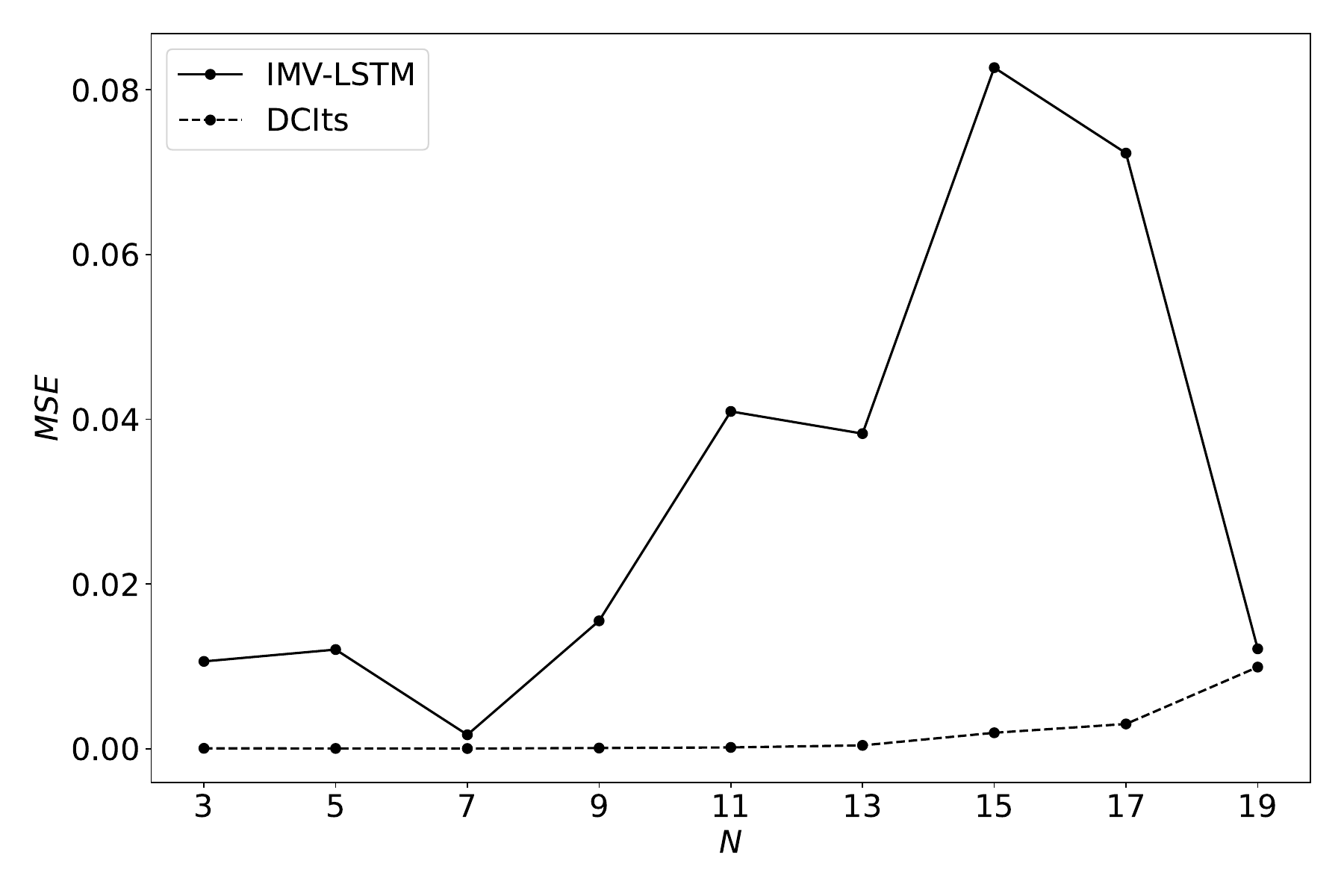}
\caption{The left figure illustrates how the stability of prediction performance for a specific model varies with noise frequency $f$ for Dataset 2. DCIts is more resilient to changes in noise frequency, with increasing instability only at the highest noise frequencies. The right figure examines the stability of prediction performance as the number of time series $N$ increases in Dataset 2. Here, DCIts again demonstrates greater stability compared to IMV-LSTM across all values of $N$.}
\label{Stress test}
\end{figure}

The stability of prediction performance was assessed on Dataset 2 with parameters ranging as follows:
\begin{itemize}
\item Noise frequency $f$ varied from 0 to 1 with a step of 0.05
\item Number of time series $N$ ranged from 3 to 19 with a step of 2
\item Noise amount: $\sigma^2_{\mathrm{noise}}$---values are: 0.01, 0.05, 0.1, 0.2, 0.5, 1, 2, 3, and~5.
\end{itemize}
Figure \ref{Stress test} provides two important insights into the models' performance. On the left side of the figure, the focus is on how prediction stability varies with changes in noise frequency, represented by $f$. DCIts shows resilience against fluctuations in noise levels, only exhibiting a small increase in instability at very high frequencies. In contrast, IMV-LSTM displays a marked rise in instability for $f>0.4$. The right panel of the figure analyzes the impact of varying the number of time series, $N$, on prediction stability. In this context, DCIts shows greater stability across diverse testing conditions than IMV-LSTM. Additionally, both models demonstrate similar behavior and achieve comparable performance for noise levels $\sigma^2_{\text{noise}} \leq 1$. However, for $\sigma^2_{\text{noise}} > 1$, DCIts achieves a significantly lower mean squared error (MSE), outperforming IMV-LSTM by an order of magnitude.

Within this controlled benchmark setting, DCIts exhibits predictive performance comparable to or better than IMV-LSTM across several stress-test regimes. As discussed in the following section, DCIts additionally exposes signed, lag-resolved coefficient tensors, allowing direct evaluation of source--lag interaction recovery on datasets with known ground truth. Dataset~2 was used for the displayed sensitivity analysis because it isolates the effects of noise frequency, noise amplitude, and dimensionality in a controlled autoregressive setting. We also repeated the high-noise robustness check on Dataset~7, which contains signed autoregressive and cross-series couplings, bias terms, and multiple relevant lags. When the noise-amount parameter $\sigma^2_{\mathrm{noise}}$ was increased from $0.1$ to $3.0$, with noise present at every time step, the MSE increased consistently with the imposed noise amount. However, the recovered nonzero coefficients remained stable: for example, $\alpgt_{2,1,2}=-1$ was recovered as $-0.973\pm0.008$ and $-1.00\pm0.02$, while $\alpgt_{4,5,1}=-0.7143$ was recovered as $-0.692\pm0.006$ and $-0.70\pm0.01$ for the two noise levels. Thus, the interpretable coefficients remain robust in this controlled perturbation.

The following qualitative analyses therefore focus directly on coefficient recovery, where the synthetic benchmarks provide known ground-truth source series, lags, signs, and coefficient values.

\subsection{Optimal window size examples}\label{optimal-wt-ex}

\begin{figure}[htb]
\centering
\includegraphics[width=12 cm]{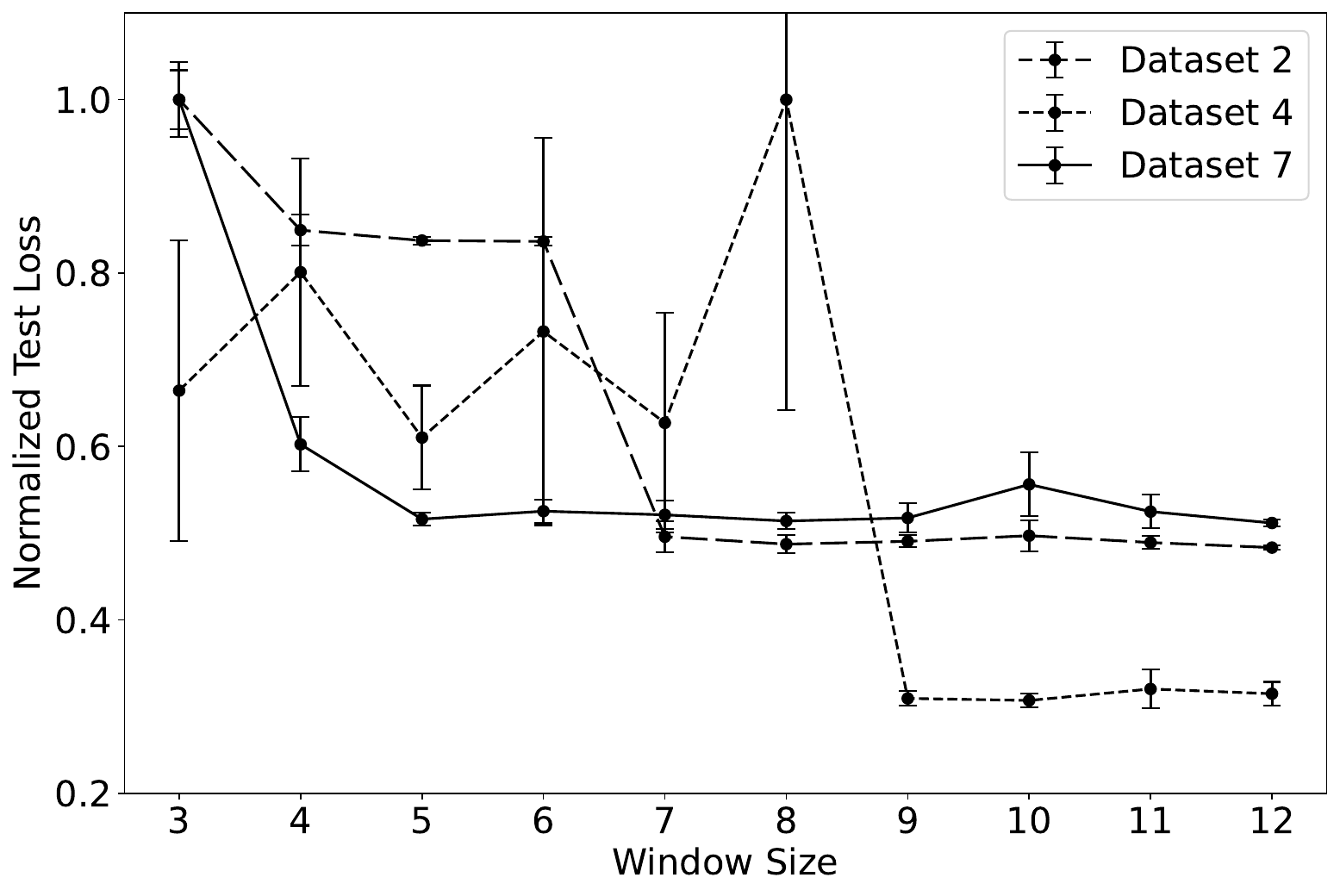}
\caption{Impact of window size on loss. On the x-axis, we plot window size, on the y-axis, we plot normalized loss on the test set. Optimal window sizes for Datasets 2, 4, and 7 are 7, 9, and 5, respectively. We observe a pronounced drop in the loss near the optimal window size and low variability of the loss across runs in that region.}
\label{Loss vs. L}
\end{figure}

We present optimal window size results for Datasets 2, 4, and 7, with $R=5$, $L_{\mathrm{min}}=3$, $L_{\mathrm{max}}=12$, and $L_{\mathrm{step}}=1$.  We expect to achieve the best results for $L_\text{o}=7$ for Dataset 2, $L_\text{o}=9$ for Dataset 4, and $L_\text{o}=5$ for Dataset 7. Figure \ref{Loss vs. L} shows the normalized test loss (using MAE as loss function) by window size, and we can see that, for optimal window sizes, we have a significant drop in loss function and low variability of the loss function in positions that correspond to the optimal window.

\begin{figure}[htb]
\centering
\includegraphics[width=9 cm]{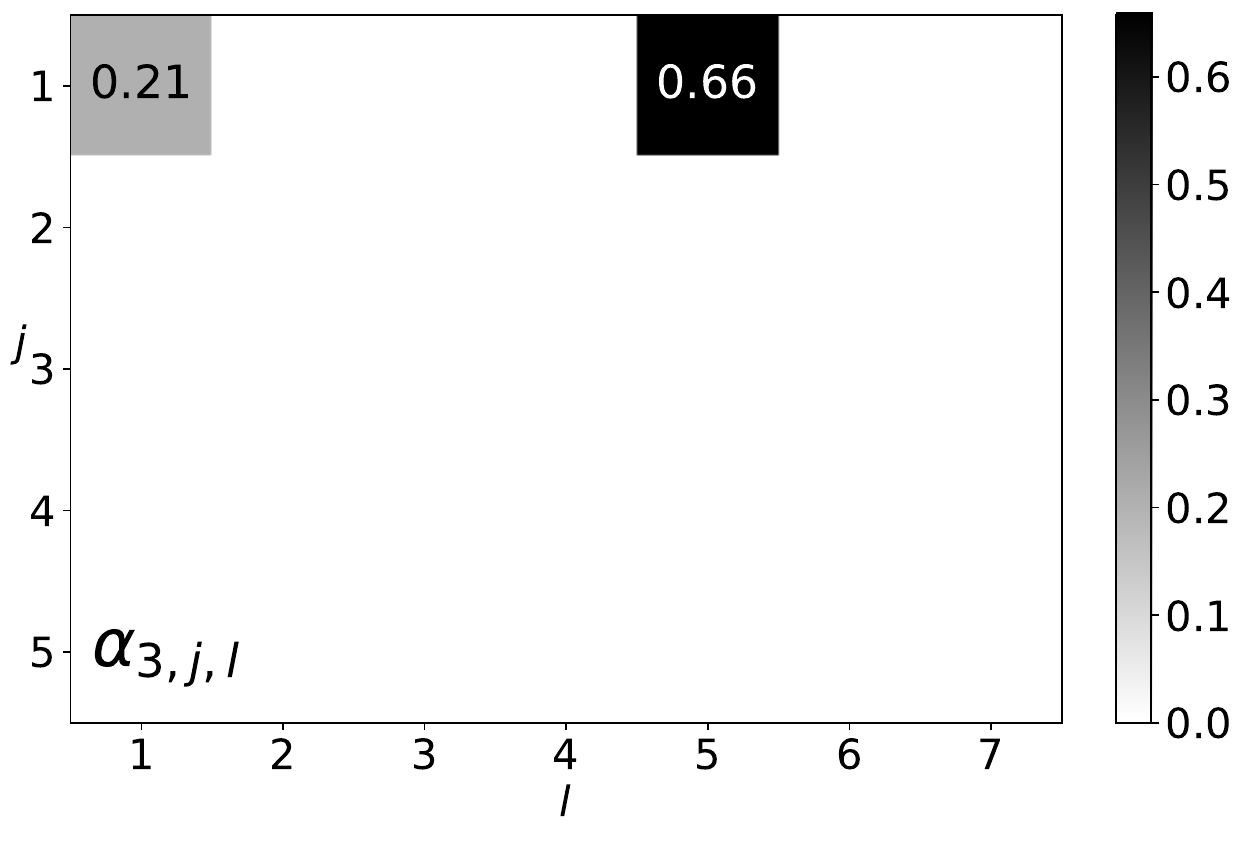}
\caption{Only coefficients passing the stability filter (Eq.~\eqref{eq:stability_mask}) are shown. $\alpha_{3,j,\ell}$ values for Dataset~7 with $L=7$ and $R=5$. No stable $\alpha_{3,j,\ell}$ entries with $\ell>5$ are retained by the stability filter.}
    \label{lmax}
\end{figure}

As previously noted, DCIts demonstrates robustness in accurately identifying interactions even when a window size larger than optimal is selected. Figure~\ref{lmax} presents the robust (run-consistent) $\alpha_{3,j,\ell}$ values for Dataset~7 with a window size of $L=7$ and $R=5$ runs. Specifically, the mean $\alpha$ coefficients were computed on the test set across five runs, and only elements that were robust (run-consistent) and non-zero were retained. Notably, in the displayed slice, all $\alpha_{3,j,\ell}$ entries with $\ell>5$ were nonsignificant.

\subsection{Qualitative analysis}\label{quali-analysis}

Next, we examine the interpretability and faithfulness of the results produced by the DCIts architecture. Before presenting the results for the benchmarking datasets, we first use a simple second-order vector autoregressive (VAR(2)) process as a transparent reference case for coefficient recovery and coefficient stability, not as an additional deep-learning baseline. Because its ground-truth linear transition matrices are known exactly, it provides a simple sanity check for whether the learned DCIts coefficients can reproduce signed lag-resolved interactions in a setting where the correct answer is unambiguous. We emphasize that, for systems adequately described by a low-order global linear VAR model, VAR remains the simpler and preferable model; the role of DCIts is to extend coefficient-level interpretability to settings where the effective interactions may be nonlinear, sample-specific, or regime-dependent:
$$
\bm{X}_t = \bm{A}_1\bm{X}_{t-1} + \bm{A}_2\bm{X}_{t-2} + \bm{\epsilon}_t
$$
where $\bm{X}^T_t = (X_{1,t}, X_{2,t}, X_{3,t})$, and the lag-1 and lag-2 matrices $\bm{A}_1$ and $\bm{A}_2$ are of dimension $3 \times 3$. The noise terms $\bm{\epsilon}_t$ are sampled from a Gaussian distribution $\mathcal{N}(0,0.2)$. 

The VAR(2) process therefore serves as a minimal transparent test of coefficient recovery. It allows us to examine whether DCIts reconstructs the known lagged interaction matrices on held-out test-set windows, using the sample-specific transition tensors reported by the model, and whether the recovered coefficients are stable across both samples and repeated training runs.

The identified optimal window size is $L_\text{o} = 2$. Using the $\bm{\alpha}$ coefficients, we can easily reconstruct the coefficient matrices used in generating the time series. Specifically, we select the $\bm{\alpha}$ coefficients corresponding to lag-1 and lag-2. For the VAR(2) process, the matrices $\bm{A}_1$ and $\bm{A}_2$ reconstructed by DCIts are presented below. The ground truth values are shown in bold font, and beneath each ground truth value, we present the DCIts reconstruction as the mean $\pm$ standard deviation of the $\alpha_{i,j,l}$ coefficients. The mean value is rounded to 2 decimal places for presentation purposes, but the results show that all significant digits are zero after the second decimal place.
$$
\renewcommand{\arraystretch}{0.7}
A_{1} = \begin{pmatrix}
\begin{array}{c}
\mathbf{0.40} \\
0.40 \pm 10^{-5}
\end{array} & 
\begin{array}{c}
\mathbf{0.10} \\
0.10 \pm 10^{-4}
\end{array} & 
\begin{array}{c}
\mathbf{0.05} \\
0.05 \pm 7 \cdot 10^{-5}
\end{array} \\[0.3cm]
\begin{array}{c}
\mathbf{0.10} \\
0.10 \pm 2 \cdot 10^{-5}
\end{array} & 
\begin{array}{c}
\mathbf{0.40} \\
0.40 \pm 2 \cdot 10^{-4}
\end{array} & 
\begin{array}{c}
\mathbf{0.10} \\
0.10 \pm 10^{-4}
\end{array} \\[0.3cm]
\begin{array}{c}
\mathbf{0.05} \\
0.05 \pm 5 \cdot 10^{-6}
\end{array} & 
\begin{array}{c}
\mathbf{0.02} \\
0.02 \pm 5 \cdot 10^{-5}
\end{array} & 
\begin{array}{c}
\mathbf{0.40} \\
0.40 \pm 5 \cdot 10^{-5}
\end{array}
\end{pmatrix}
\renewcommand{\arraystretch}{1}
$$
$$
\renewcommand{\arraystretch}{0.7}
A_{2} = \begin{pmatrix}
\begin{array}{c}
\mathbf{0.20} \\
0.20 \pm 10^{-4}
\end{array} & 
\begin{array}{c}
\mathbf{0.05} \\
0.05 \pm 10^{-5}
\end{array} & 
\begin{array}{c}
\mathbf{0.02} \\
0.02 \pm 4 \cdot 10^{-6}
\end{array} \\[0.3cm]
\begin{array}{c}
\mathbf{0.05} \\
0.05 \pm 10^{-5}
\end{array} & 
\begin{array}{c}
\mathbf{0.20} \\
0.20 \pm 2 \cdot 10^{-4}
\end{array} & 
\begin{array}{c}
\mathbf{0.05} \\
0.05 \pm 5 \cdot 10^{-6}
\end{array} \\[0.3cm]
\begin{array}{c}
\mathbf{0.02} \\
0.02 \pm 5 \cdot 10^{-5}
\end{array} & 
\begin{array}{c}
\mathbf{0.05} \\
0.05 \pm 7 \cdot 10^{-5}
\end{array} & 
\begin{array}{c}
\mathbf{0.20} \\
0.20 \pm 10^{-4}
\end{array}
\end{pmatrix}
\renewcommand{\arraystretch}{1}
$$

Our discussion primarily focuses on the interpretability provided by DCIts for the datasets listed in Appendix~\ref{bdata}. Unless explicitly stated otherwise, these qualitative analyses use the default bias-plus-linear DCIts configuration with active orders $p=0$ and $p=1$. The optimal window size has been discussed in Section~\ref{optimal-wt-ex}, and we use it for the remaining benchmarking datasets. All reported $\bm{\alpha}$ and $\bm{\beta}$ coefficients are based on mean values across five runs, with bias terms reported only when they are stable and non-negligible. This repeated-run procedure assesses the stability of the recovered coefficients under different random initializations while using a fixed synthetic realization for the time series.

\subsubsection{Autoregressive Datasets}\label{sec:auto}

\begin{figure}[htb]
\centering
\includegraphics[width=5.3 cm]{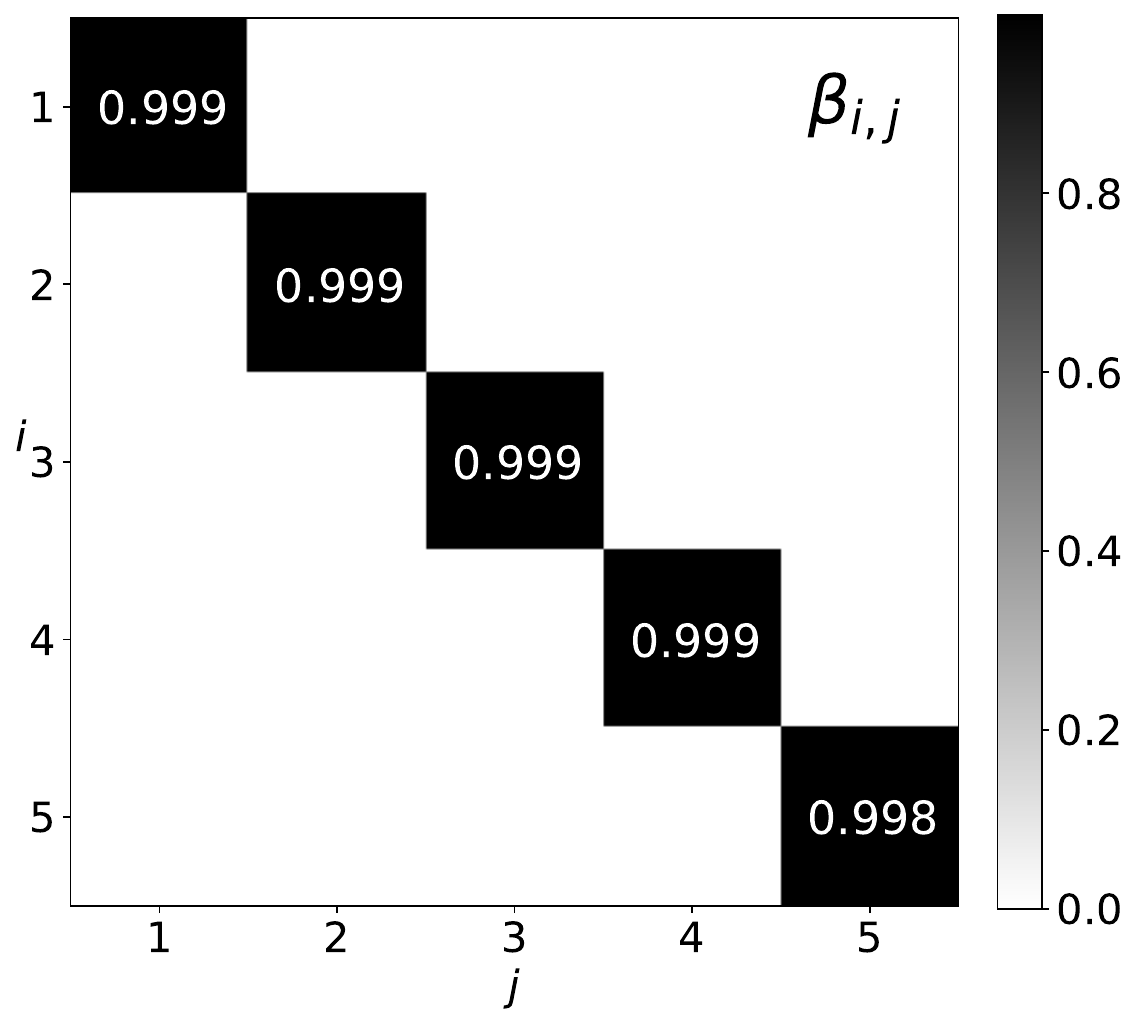}
\includegraphics[width=7 cm]{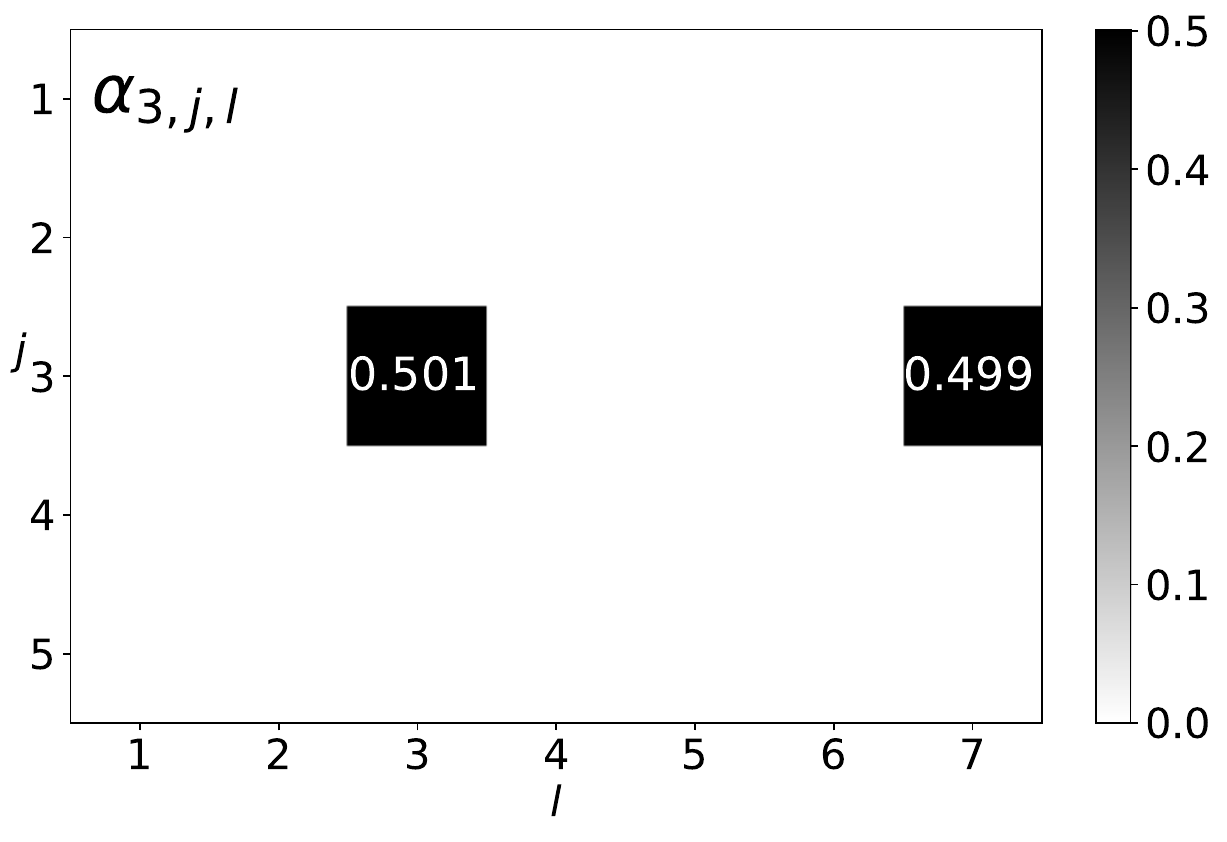}
\caption{In the left panel, the $\bm{\beta}$ coefficients estimated by DCIts for Dataset 2 are shown. The y-axis corresponds to the target series index, and the x-axis corresponds to the source time series. Since this dataset consists solely of autoregressive time series with no interactions, the ground truth $\bm{\beta}$ values correspond to the identity matrix, with nonzero values only on the diagonal. The model correctly identifies these dynamics, producing near-zero off-diagonal and diagonal elements close to one. In the right panel, $\bm{\alpha}$ coefficients are shown for a single time series $X_{3,t}$. The y-axis corresponds to the time series index ($j$), and the x-axis corresponds to the lag ($l$) of $\alpha_{3,j,l}$ coefficient. Lags 3 and 7 are the key contributors, closely matching the ground-truth value of 0.5. All other $\alpha_{3,j,l}$ values are significantly smaller.}
\label{Interpretability dataset 2}
\end{figure}

We will start by analyzing the interpretability and faithfulness of the DCIts architecture on benchmarking datasets using Dataset 2. For brevity, detailed results for Dataset 1 are omitted here. In Dataset 1, each time series is represented by a constant, and DCIts accurately identifies the generating process of each time series. The corresponding results can be found in the examples folder of \cite{githubdcits}. Dataset 2 consists of five autoregressive time series that do not interact with each other. In the left panel of Figure~\ref{Interpretability dataset 2}, the $\bm{\beta}$ coefficients estimated by DCIts are shown. Since the dataset consists solely of autoregressive time series, the ground truth for the $\bm{\beta}$ coefficients is the identity matrix, as determined by the generating $\bm{\alpha}$ values. As illustrated in the left panel of Figure \ref{Interpretability dataset 2}, the model successfully captures the underlying dynamics of Dataset 2, producing correct interpretability with high confidence. The mean values of the $\bm{\beta}$ diagonal elements across all samples range from 0.9979 to 0.9994, with standard deviations between $10^{-4}$ and $3 \cdot 10^{-4}$, the ground truth being 1. In contrast, the mean values of the off-diagonal elements range from $10^{-4}$ to $7 \cdot 10^{-4}$, with standard deviations between $10^{-4}$ and $10^{-3}$, the ground truth is 0.

In the right panel of Figure \ref{Interpretability dataset 2}, the $\alpha_{3,j,l}$ coefficients are displayed for a single time series $X_{3,t}$ in Dataset 2 (a similar pattern is observed for all other time series in this dataset). The lags 3 and 7 are the key contributors to generating the next step in the time series. DCIts correctly identifies that Dataset 2 is an autoregressive process with no interaction between the time series; that is, $X_{3,t}$ is generated solely by $X_{3,t}$. DCIts also accurately recognized the most significant lags, with the $\alpha_{3,3,l}$ coefficients for lags 3 and 7 ($\alpgt_{3,3,3}=\alpgt_{3,3,7}=0.5$) being $\alpha_{3,3,3} = (0.4992 \pm 0.0007)$ and $\alpha_{3,3,7}= (0.5010 \pm 0.0008)$, while all other $\bm{\alpha}$ values are significantly smaller and/or not run-consistent. Additionally, the $\bm{\alpha}$ coefficients exhibit low variability across all samples. 

A similar behavior is observed in Dataset 3 when using the default bias-plus-linear DCIts configuration, even though Dataset 3 is generated by a nonlinear process and is therefore also suitable for analysis with higher-order DCIts branches.

\subsubsection{Datasets with cross correlation}
Datasets 4, 5, and 6 are used to assess the performance of DCIts when cross-correlations are present. In our previous work \cite{baric2021}, these datasets presented a significant challenge to benchmarked interpretable models.

\begin{figure}[htb]
\centering
\includegraphics[width=12 cm]{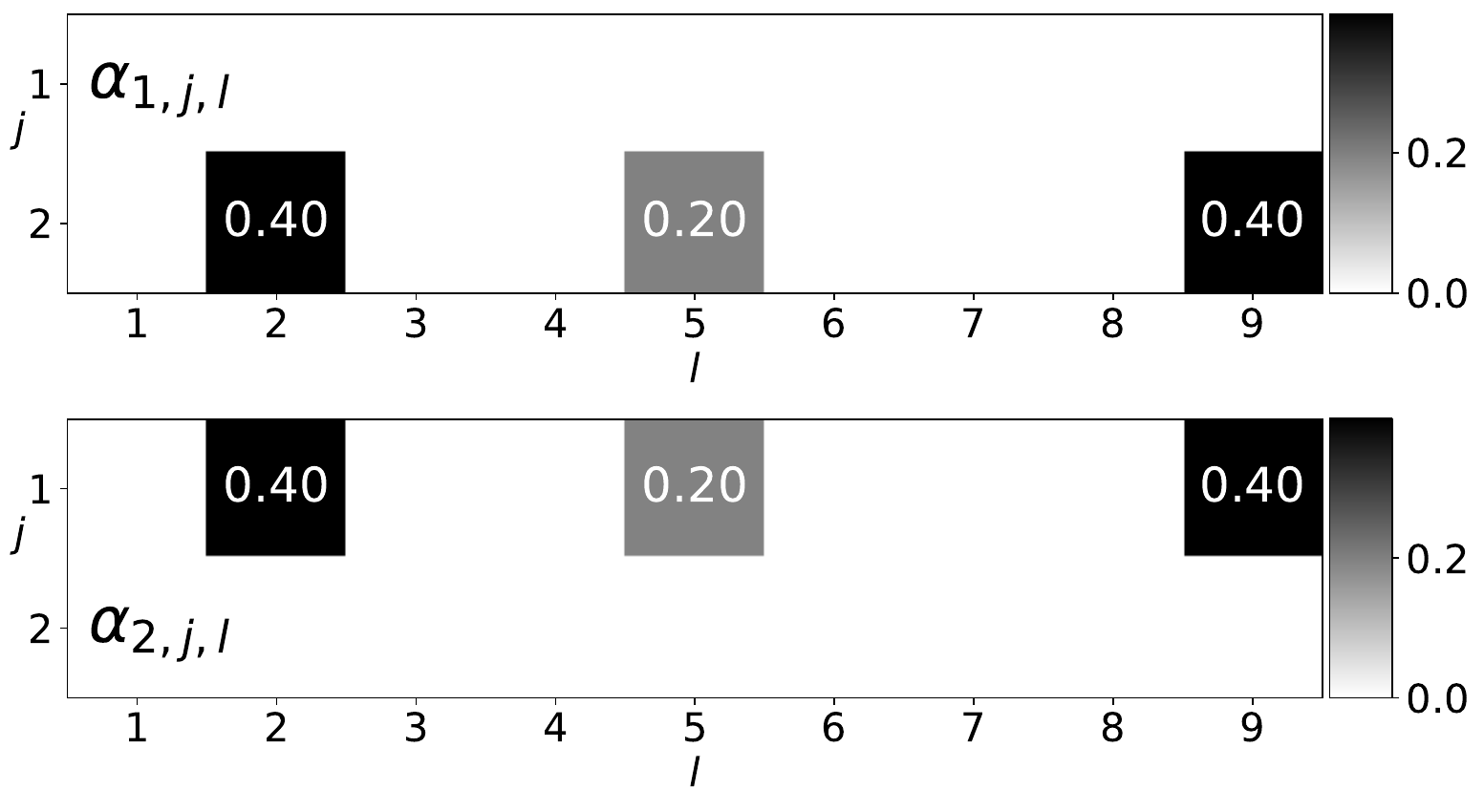}
\caption{Mean $\bm{\alpha}$ values for Dataset 4, obtained using MAE as the loss function, match ground truth values. The y-axis corresponds to the source time series index ($j$), and the x-axis represents the time lag ($l$).}
\label{Interpretability dataset 4}
\end{figure}

Dataset 4 is characterized by only cross-correlations, with no autocorrelation. When analyzing the mean and standard deviation of the $\bm{\alpha}$ coefficients obtained for Dataset 4 using the MAE loss function in Figure~\ref{Interpretability dataset 4}, it is evident that our extraction procedure successfully identified values corresponding to the ground truth. Furthermore, this procedure provides a highly stable interpretation across all samples, accurately identifying the correct interactions and lags. Specifically, values are  $\alpha_{1,2,2} = (0.396 \pm 0.004)$, $\alpgt_{1,2,2} = 2/5$,  $\alpha_{1,2,5} = (0.197 \pm 0.005)$, $\alpgt_{1,2,5} = 1/5$, $\alpha_{1,2,9} = (0.398 \pm 0.004)$, $\alpgt_{1,2,9} = 2/5$, $\alpha_{2,1,2} = (0.399 \pm 0.006)$, $\alpgt_{2,1,2} = 2/5$, $\alpha_{2,1,5} = (0.196 \pm 0.004)$, $\alpgt_{2,1,5} = 1/5$, $\alpha_{2,1,9} = (0.397 \pm 0.008)$, $\alpgt_{2,1,9} = 2/5$. As we mentioned, interpretability is robust to the choice of loss function, but in Dataset 4, the MAE loss performs better than the MSE loss. When MSE is used as a loss function, the validation loss does not follow a smooth trend and exhibits large oscillations.

\begin{figure}[!htb]
\centering
\includegraphics[width=5 cm]{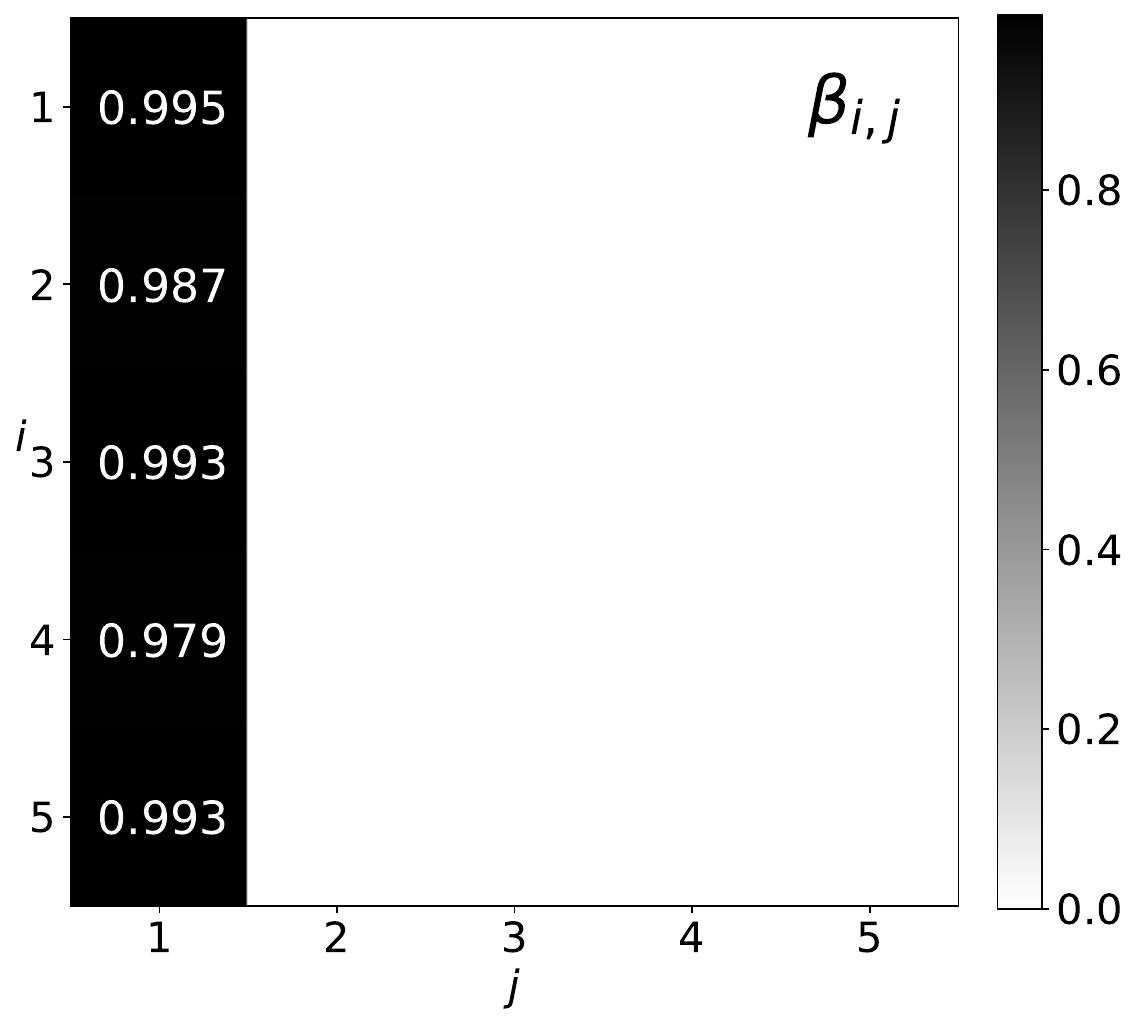}
\includegraphics[width=8cm]{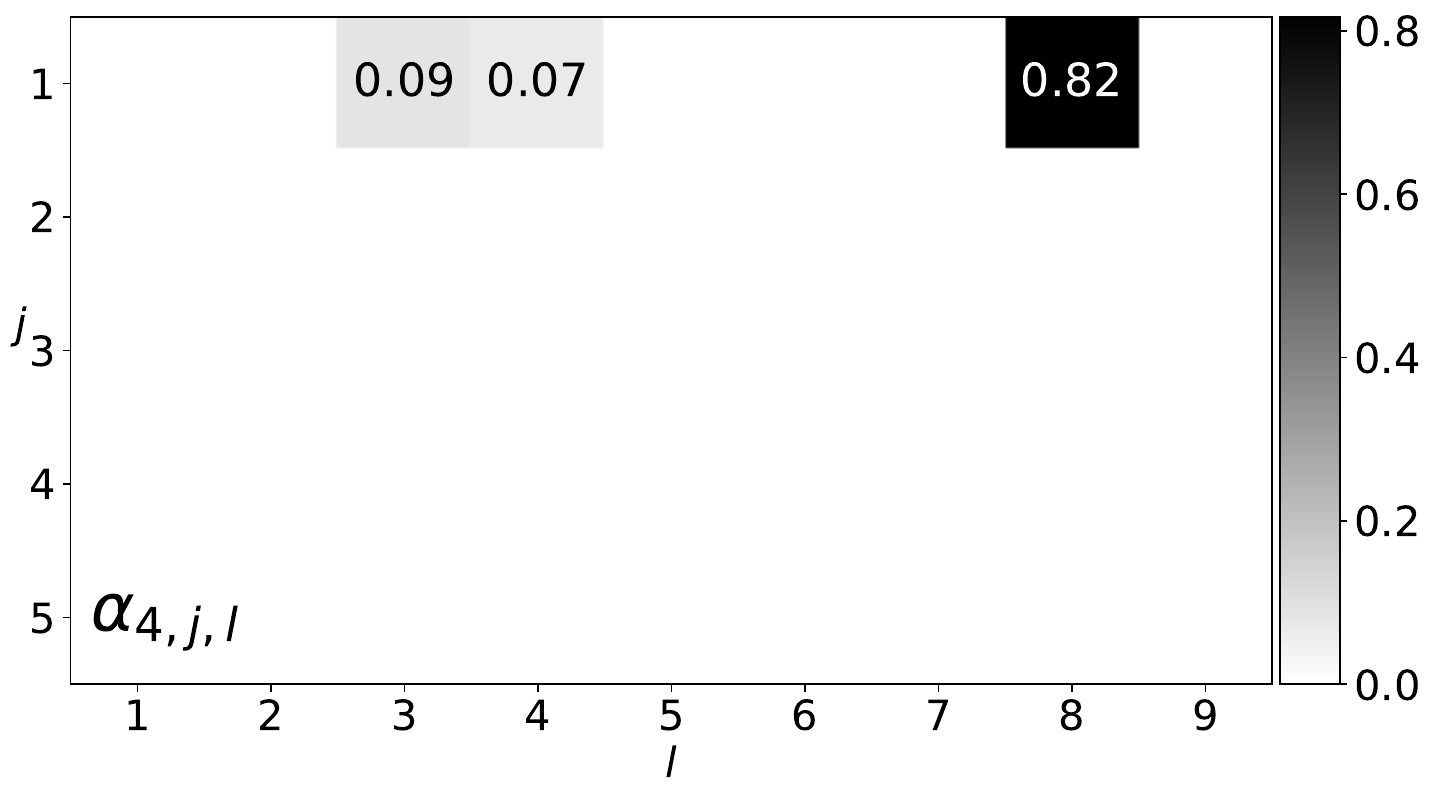}
\caption{Mean $\bm{\beta}$ values estimated by DCIts using MAE (left panel) and $\alpha_{4,j,l}$ values (right panel) for Dataset 5. The ground truth is given by $\beta_{i,1}=1$ for all $i$, with all other $\bm{\beta}$ values being zero. Each time series in this dataset depends solely on $X_{1,t}$, and DCIts correctly identified $X_{1,t}$ as a source for all time series. DCIts correctly identifies important lags with relatively smaller impact in $\alpha_{4,j,l}$.}
\label{Interpretability dataset 5}
\end{figure}

In Datasets 5 and 6, each time series is generated based on the first time series, $X_{1,t}$. Since Dataset 6 is more suited for the higher-order version of DCIts, the focus here remains on Dataset 5. In Figure \ref{Interpretability dataset 5}, we present the $\bm{\beta}$ and $\alpha_{4,j,l}$ values estimated by DCIts using MAE as the loss function for Dataset 5.

The ground truth for Dataset 5 is given by $\beta_{i,1}=1$ for all $i$, with all other $\bm{\beta}$ values being zero. The mean values of $\beta_{i,1}$ across all samples fall within (0.98, 0.99), with standard deviations ranging from 0.01 to 0.03. All other $\bm{\beta}$ values equal zero.

Analyzing interpretability at the single time-series level for Dataset 5 reveals that DCIts, using the MAE as the loss function, accurately identifies the most significant lags. For instance, in the case of $X_{5,t}$, the ground truth coefficients are $\alpgt_{5,1,2} = 1/3$, $\alpgt_{5,1,5} = 2/9$, and $\alpgt_{5,1,8} = 4/9$. The model identified all significant lags with low variability across runs, yielding  $\alpha_{5,1,2} = (0.336 \pm 0.003)$, $\alpha_{5,1,5} = (0.222 \pm 0.004)$, and $\alpha_{5,1,8} = (0.43 \pm 0.01)$. All other $\alpha_{5,j,l}$ values were correctly estimated as nonsignificant.

Similarly, for $X_{4,t}$, the ground truth coefficients are $\alpgt_{4,1,3} = 1/10$, $\alpgt_{4,1,4} = 1/10$, and $\alpgt_{4,1,8} = 4/5$. The model again successfully and with high confidence identified all significant lags, resulting in $\alpha_{4,1,3} = (0.09 \pm 0.01)$, $\alpha_{4,1,4} = (0.07 \pm 0.03)$, and $\alpha_{4,1,8} = (0.82 \pm 0.01)$. These results demonstrate DCIts's capability to accurately interpret important lags, even those with relatively smaller impacts.

\subsubsection{High mixture of behaviours}
With Dataset 7, we evaluate the performance of DCIts in scenarios where bias and a mixture of different dynamics exist between time series in the generating process, including autoregressive processes and cross-correlations that may be positive or negative. In this dataset, the time series $X_{2,t}$ and $X_{4,t}$ include constant terms in their generating processes, with ground-truth values $b^{\text{gt}}_{2} = 1$ and $b^{\text{gt}}_{4} = 1$. Using both bias and linear terms ($p = 0 \text{ and } 1$), DCIts accurately identified the bias values as $b_{2} = (0.89 \pm 0.07)$ and $b_{4} = (0.97 \pm 0.02)$.

\begin{figure}[!htb]
\centering
\includegraphics[width=6 cm]{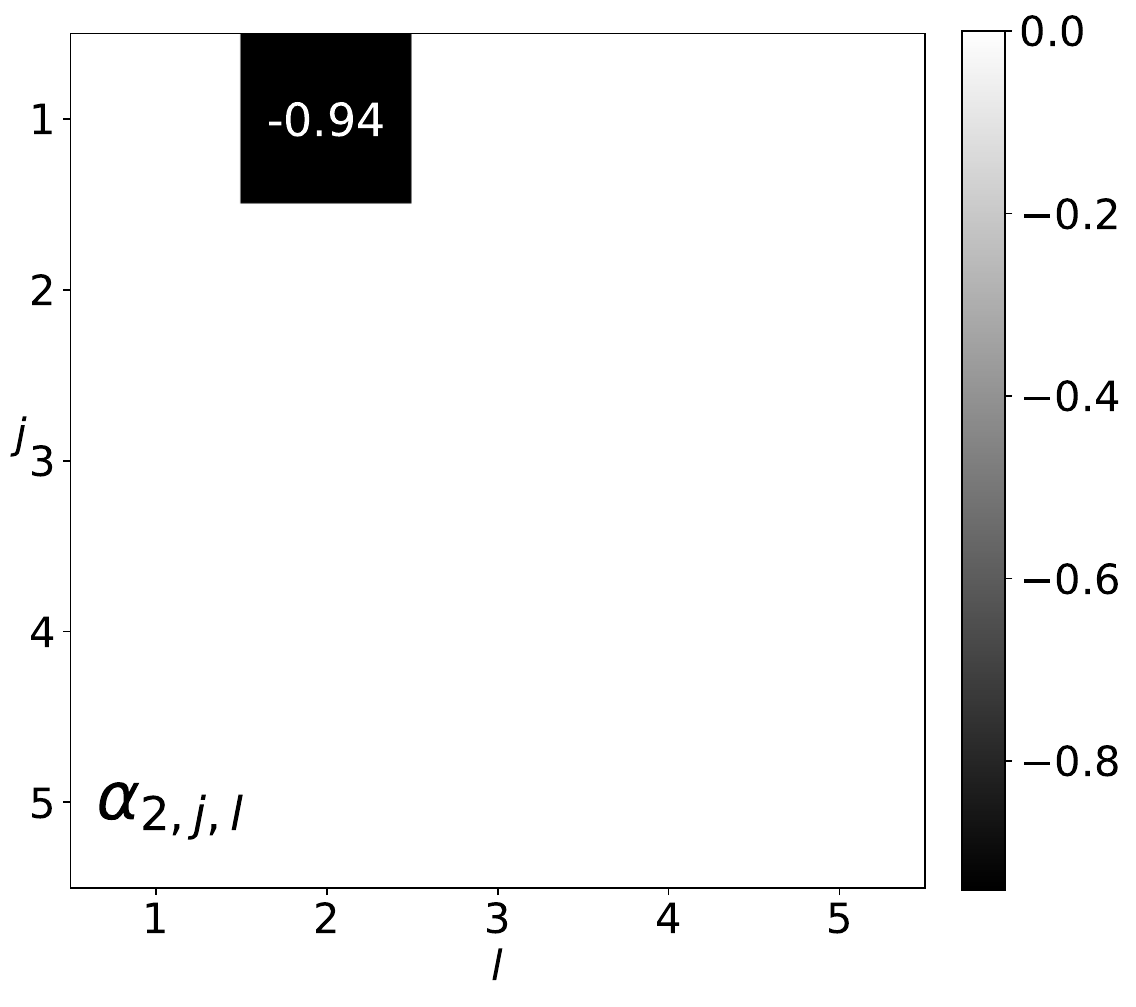}
\includegraphics[width=6 cm]{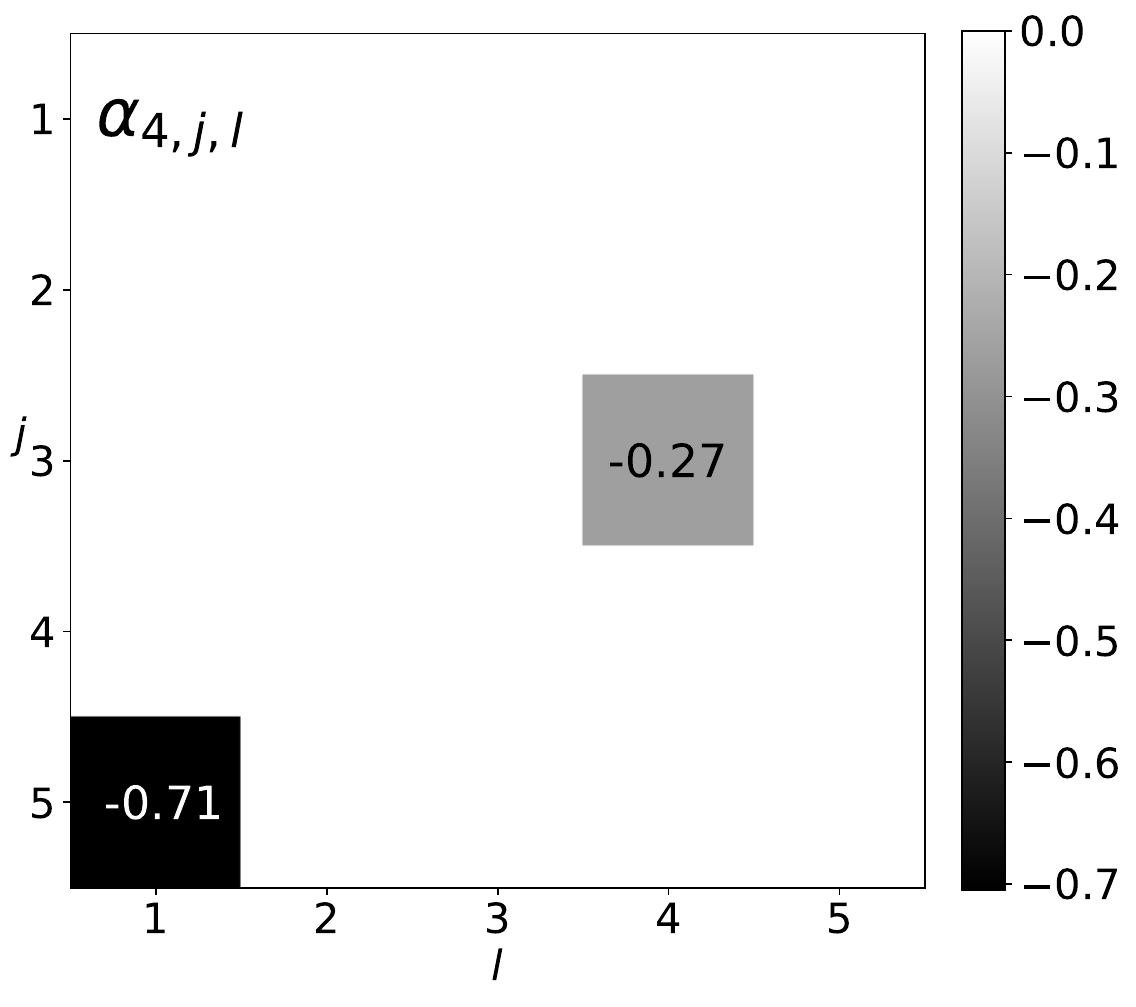}
\caption{The mean values of $\alpha_{2,j,l}$ (left panel) and $\alpha_{4,j,l}$ (right panel) for Dataset 7. The y-axis corresponds to the source time series index ($j$), and the x-axis represents the time lag ($l$). DCIts not only identifies the important time series and lags but can also detect the correct sign and strength of the interaction. For Dataset 7, DCIts identifies the source time series together with the corresponding lag, sign, and interaction magnitude.}
\label{Interpretability dataset 7}
\end{figure}

Figure \ref{Interpretability dataset 7} presents the results for the $\alpha_{2,j,l}$ and $\alpha_{4,j,l}$ coefficients: $\alpgt_{2,1,2} = -1$, $\alpha_{2,1,2} = (-0.94 \pm 0.04)$; $\alpgt_{4,3,4} = -2/7 \approx -0.286$, $\alpha_{4,3,4} = (-0.27 \pm 0.02)$; $\alpgt_{4,5,1} = -5/7 \approx -0.714$, $\alpha_{4,5,1} = (-0.71 \pm 0.01)$. DCIts successfully identified the correct lags and accurately determined the sign of the interactions.
 
\subsubsection{Time series with multiple dynamics} \label{subsubsec-Switch}
Dataset 8 is an example of a dataset that exhibits two distinct dynamics, with the active dynamic determined by $X_{1,t-5}$. This dataset allows us to test whether DCIts can detect the presence of two different dynamics.

In Dataset~8, the first series acts as a switching driver. It alternates between two bias levels, $b_{\mathrm{low}}=0.2$ and $b_{\mathrm{high}}=0.7$, with stochastic persistence times. The active regime is determined by $X_{1,t-5}$ and changes the source--lag structure for $X_{2,t}$ and $X_{3,t}$, while $X_{4,t}$ follows the same autoregressive dynamics in both regimes. The full generating equations are given in Appendix~\ref{bdata}.

Before analyzing whether DCIts can identify changes in the dataset's dynamics, we first examine the results for $X_{4,t}$ to assess whether the model correctly captures the generating process for time series with fixed dynamics. $\alpha_{4,j,l}$ values are $\alpha_{4,4,1} = (0.50001 \pm 0.0001)$, and $\alpha_{4,4,4} = (0.40001 \pm 0.0003)$, for $X_{1,t-5} \leq 0.5$, and $\alpha_{4,4,1} = (0.4999 \pm 0.0001)$, and $\alpha_{4,4,4} = (0.4002 \pm 0.0006)$, for $X_{1,t-5} > 0.5$. The corresponding ground-truth values are $1/2$ and $2/5$, respectively. So in the case of fixed dynamics, DCIts reconstructed the generating process successfully.

Results for $X_{1,t}$ are the only instances in which DCIts deviates from perfect faithfulness. For $X_{1,t-5} \leq 0.5$, it correctly determines that there is bias present with $b_1=(0.196 \pm 0.004)$ (ground truth is 0.2), and the only non-zero $\alpha_{1,j,l}$ element is $\alpha_{1,1,1}=(0.06 \pm 0.02)$. The ground truth for all $\alpha_{1,j,l}$ values is zero. Therefore, we can conclude that the model correctly deciphered the dynamics. For $X_{1,t-5} > 0.5$, DCIts does not detect the shift in bias, remaining fixed at $b_1=(0.196 \pm 0.004)$ (ground truth 0.7), and instead utilizes $\alpha_{1,1,1}=(0.49 \pm 0.04)$. We attempted to modify the structure of the $p=0$ focuser so that the input is $Q_{t-1}$ rather than $I$, but this did not alter the outcome.

\begin{figure}[!htb]
\centering
\includegraphics[width=13.5 cm]{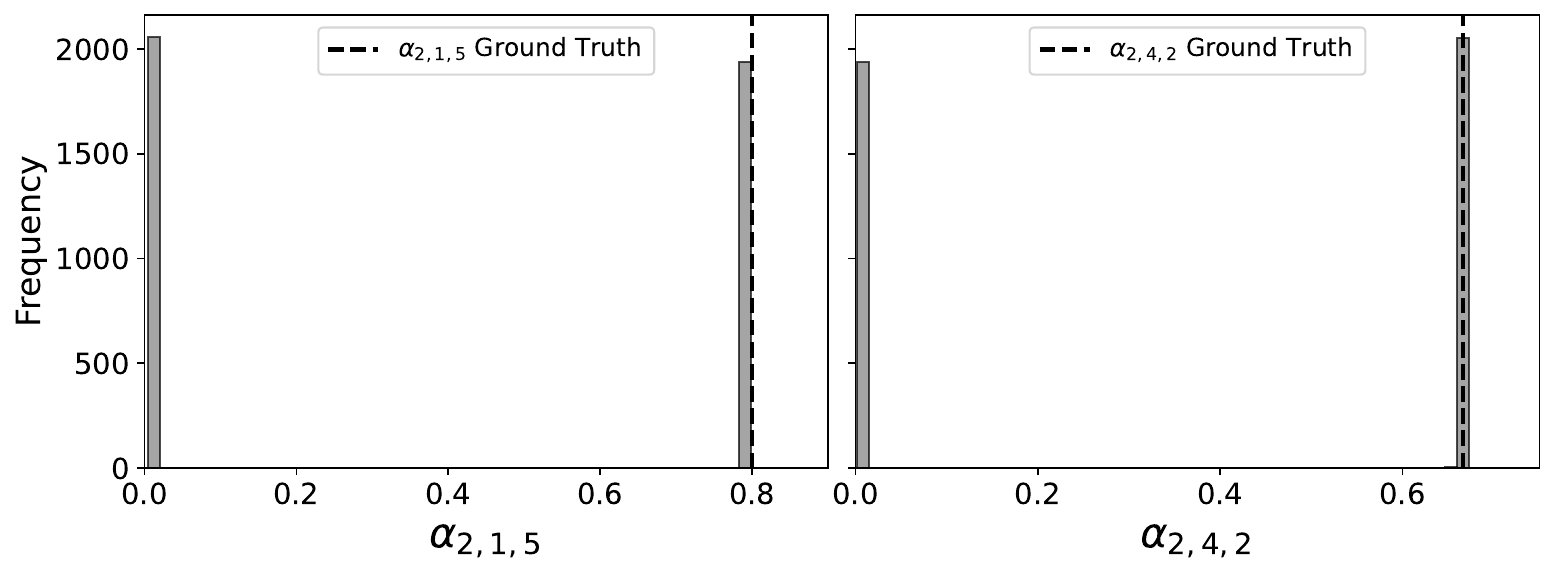}
\caption{Histograms for $\alpha_{2,1,5}$ (left) and $\alpha_{2,4,2}$ (right). Both distributions are bimodal (elements are either close to zero or to ground truth), indicating that DCIts successfully detected the two different dynamics in this dataset.}
\label{d8_interpretability_a2}
\end{figure}

\begin{figure}[!htb]
\centering
\includegraphics[width=6 cm]{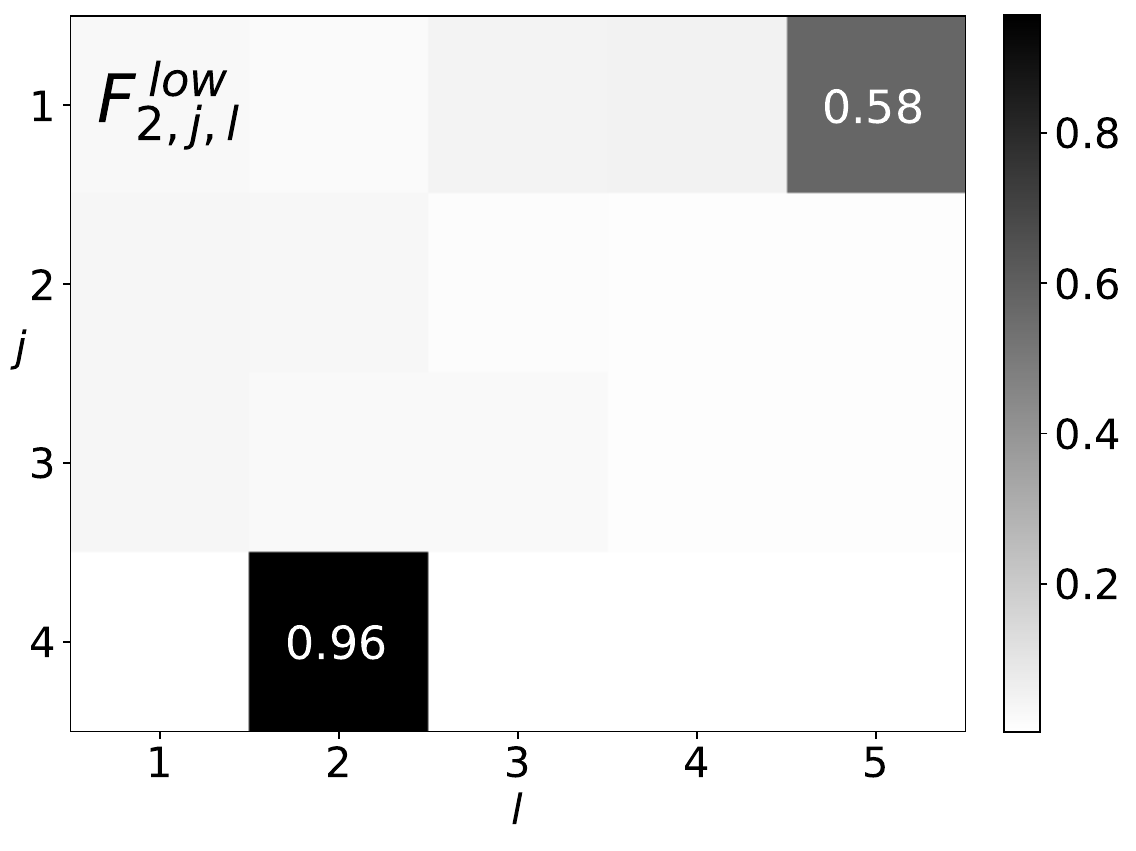}
\includegraphics[width=6 cm]{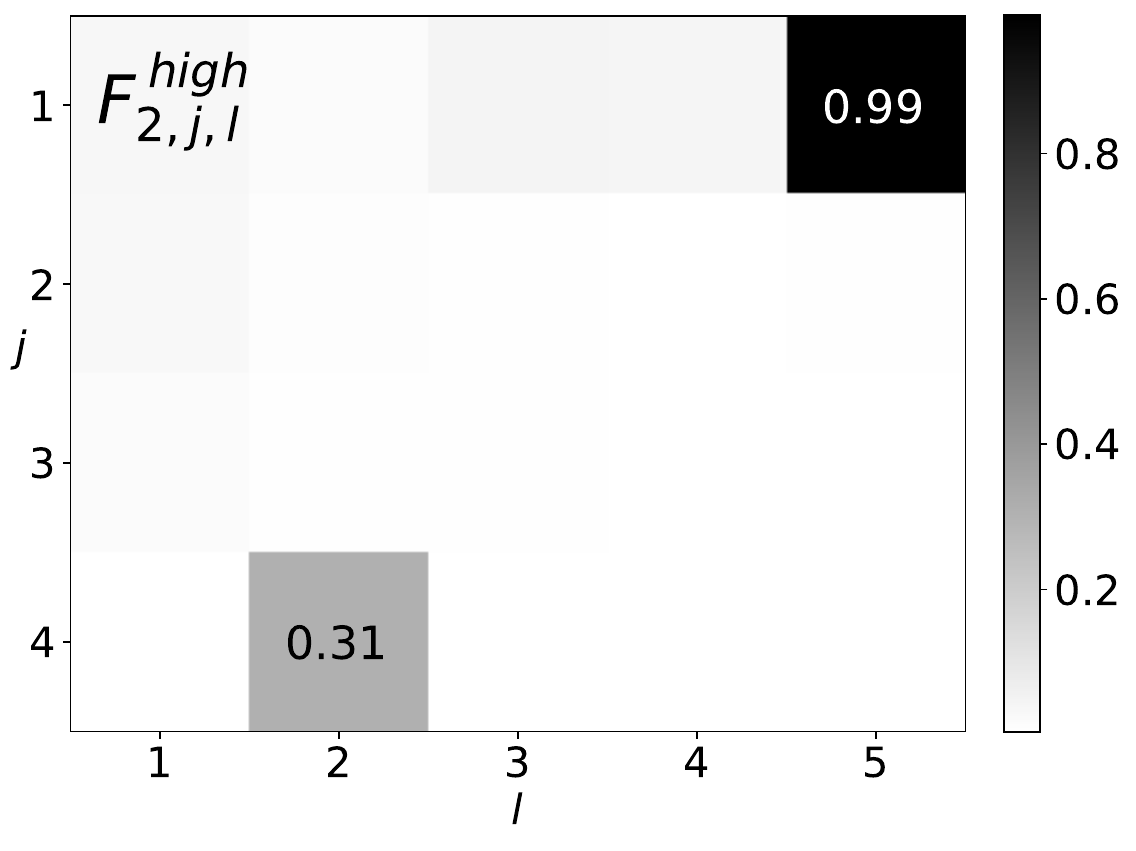}
\caption{Mean focuser $F_{2,j,l}$ values for $X_{1,t-5} \leq 0.5$ (left panel) and $X_{1,t-5} > 0.5$ (right panel) from the DCIts model for Dataset 8. The y-axis corresponds to the source time series $j$, and the x-axis corresponds to the lag $l$. The focuser correctly identified the most important lags in both dynamics, giving lower weights to the non-present generating element.}
\label{d8_interpretability_f}
\end{figure}

According to the generating process for Dataset 8, $\alpha_{2,1,5}$ is expected to be $\alpgt_{2,1,5} = 4/5$ when $X_{1,t-5} > 0.5$, and it should be zero when $X_{1,t-5} \leq 0.5$. Conversely, for $\alpha_{2,4,2}$, $\alpgt_{2,4,2}$ is expected to be zero when $X_{1,t-5} > 0.5$, and $2/3$ when $X_{1,t-5} \leq 0.5$. In Figure~\ref{d8_interpretability_a2}, we plot histograms for the values of $\alpha_{2,1,5}$ and $\alpha_{2,4,2}$ based on the value of $X_{1,t-5}$. We observe a bimodal distribution in both histograms, indicating that DCIts successfully detected two distinct dynamics in this dataset, with $\alpha_{2,4,2} = (0.6663 \pm 0.0005)$ and $\alpha_{2,1,5} = (0.796 \pm 0.005)$. The same is true for $\alpha_{3,4,4} = (0.800 \pm 0.001)$ and $\alpha_{3,1,4} = (0.665 \pm 0.002)$. For $X_{2,t}$ and $X_{3,t}$, DCIts accurately recovered the regime-dependent source--lag structure in this challenging switching case.
Figure~\ref{d8_interpretability_f} shows $F_{2,j,l}$, i.e., the $p=1$ focuser values for $X_{2,t}$. We see that this ability stems from the focuser identifying the correct time series and lag in generating dynamics, giving lower weights to the non-present generating elements in the $X_{1,t-5} \leq 0.5$ and $X_{1,t-5} > 0.5$ regimes. The final $\bm{\alpha}$ values are fixed by the modeler $\bm{C}$.

\subsection{Testing higher order DCIts}\label{higher-order-ex}

To analyze DCIts behavior with elementwise higher-order approximations, we utilize a process inspired by the cubic map $X_t=aX_{t-1}^3 + (1-a)X_{t-1}$ with $a=3.75$. This is the only experiment in which polynomial branches beyond the default bias-plus-linear configuration are activated. The goal is not to show that the higher-order model is universally better than the default model, but to test whether DCIts can separate linear and elementwise cubic source--lag contributions when such terms are present in the generating process. This experiment does not test mixed multiplicative terms such as $X_{i,t-\ell_1}X_{j,t-\ell_2}$:
\begin{align}\label{eq:cubic}
X_{1,t} &= (1-a)X_{1,t-3} + a X_{1,t-3}^3 + \epsilon_{1,t} \nonumber\\
X_{2,t} &= (1-a)X_{2,t-5} + a X_{2,t-5}^3 + \epsilon_{2,t} \\
X_{3,t} &= \frac{1}{2}(X_{1,t-3} + X_{2,t-5}) + \epsilon_{3,t}. \nonumber
\end{align}     

\begin{figure}[!htb]
\centering
\includegraphics[width=6 cm]{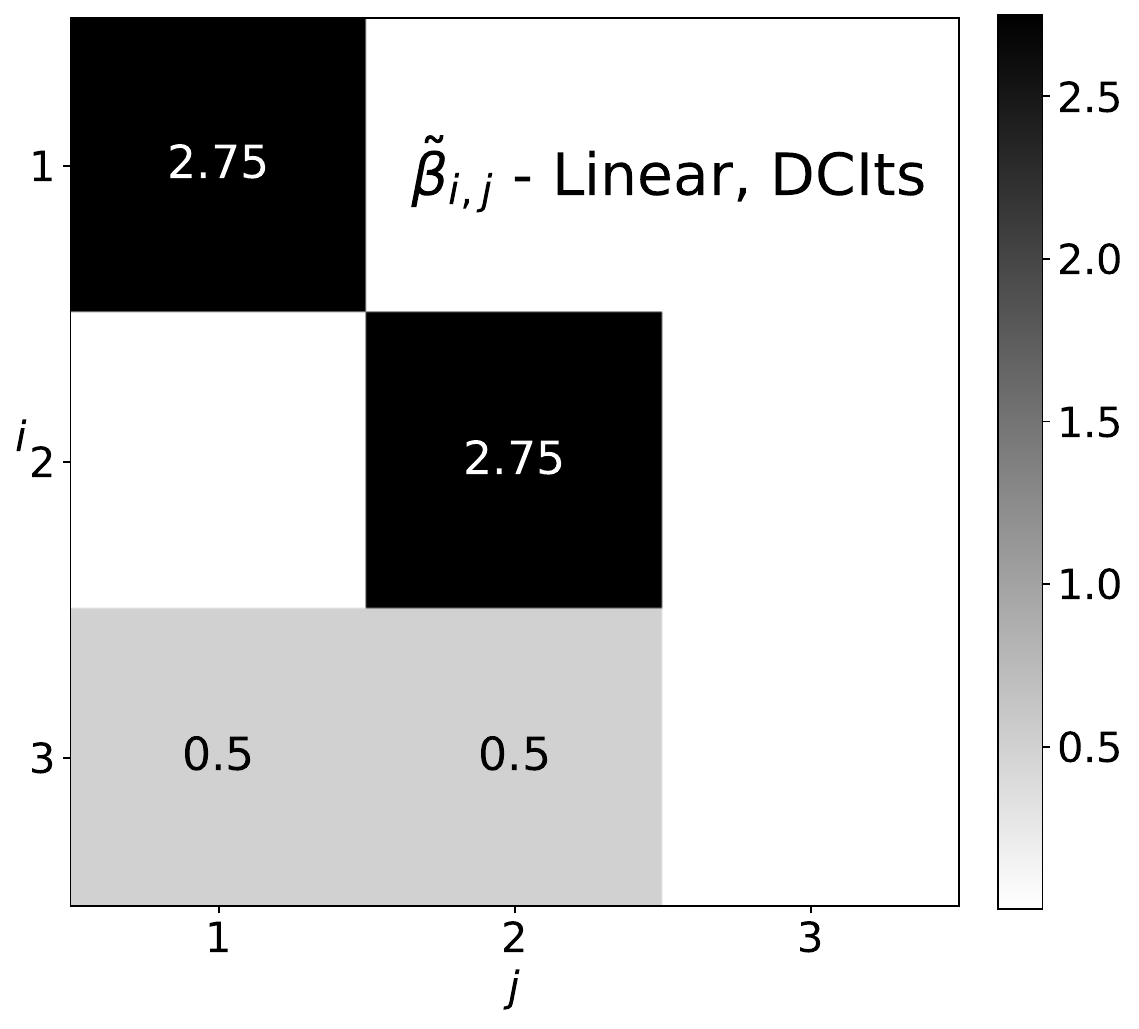}
\includegraphics[width=6 cm]{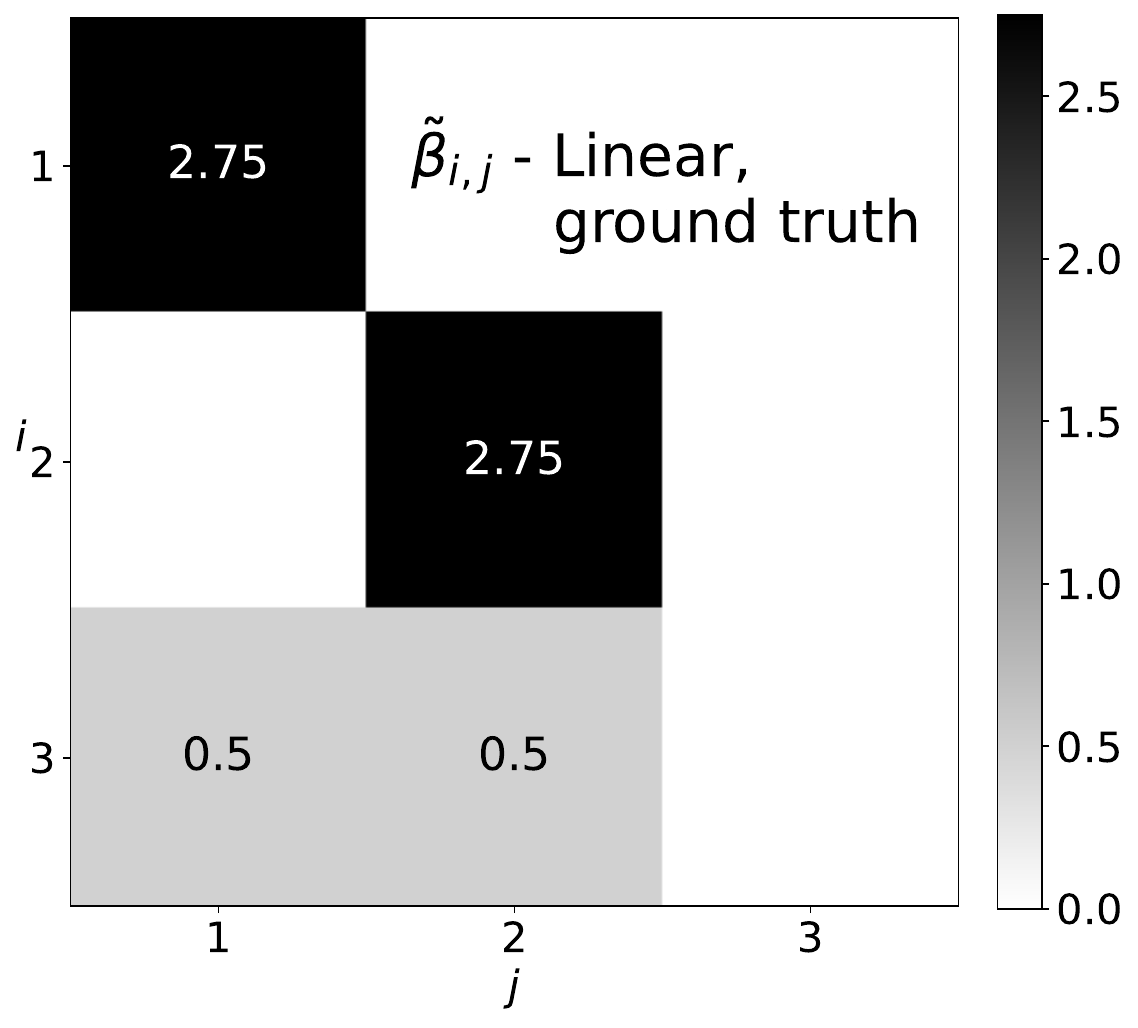}
\includegraphics[width=6 cm]{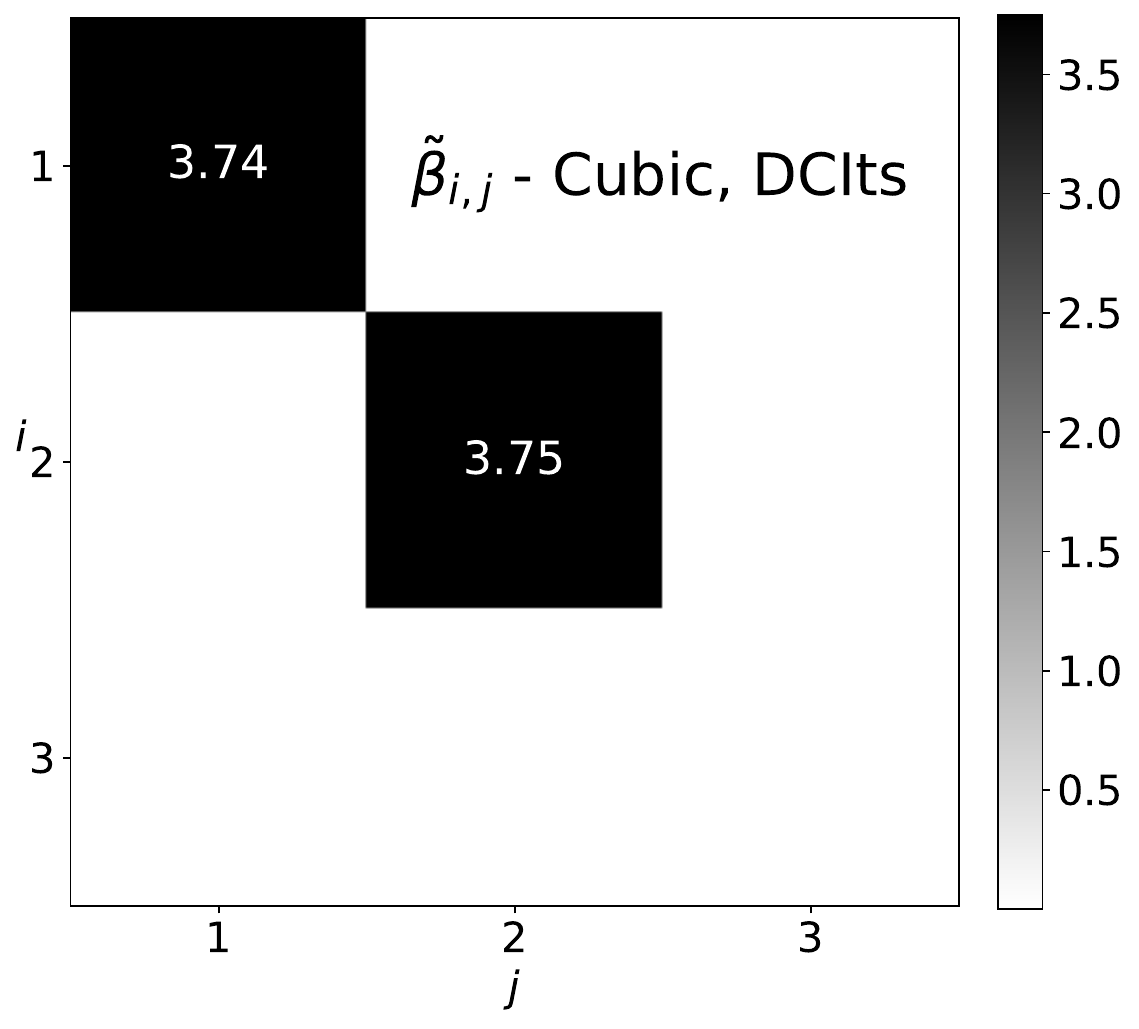}
\includegraphics[width=6 cm]{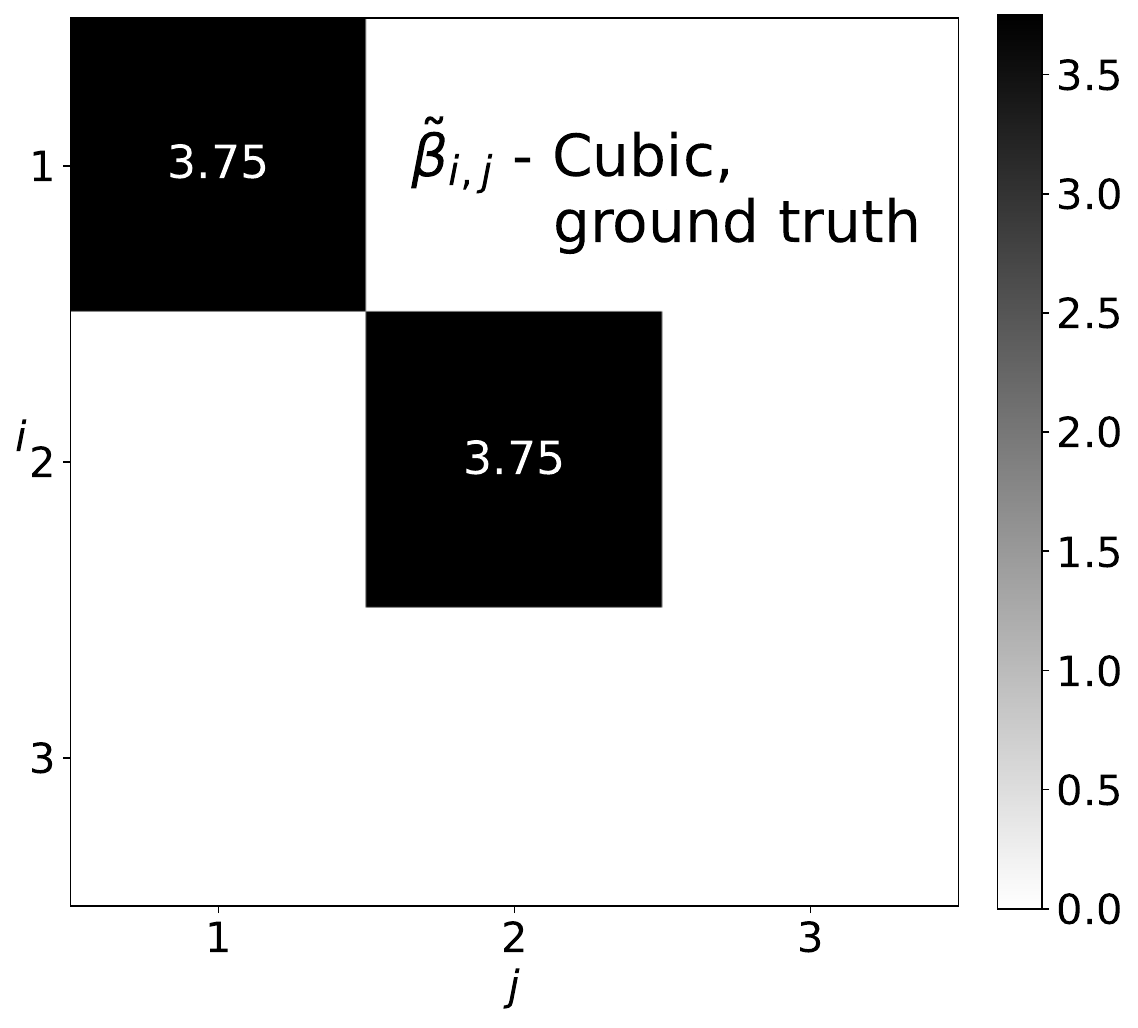}
\caption{$\tilde{\bm{\beta}}$ coefficients for the linear and cubic terms learned by DCIts, together with the corresponding ground-truth values for the cubic process. DCIts correctly recovers the linear and cubic terms for $X_{1,t}$ and $X_{2,t}$, and identifies that $X_{3,t}$ is generated by a linear combination of lagged values of $X_{1,t}$ and $X_{2,t}$, with no cubic contribution to $X_{3,t}$.}
\label{Interpretability_cubic}
\end{figure}

In Figure~\ref{Interpretability_cubic}, we plot the $\tilde{\bm{\beta}}$ values for the linear and cubic terms of DCIts and compare them to the ground truth for the Cubic process, Equation~(\ref{eq:cubic}). We use $\tilde{\bm{\beta}}$ instead of $\bm{\beta}$ to emphasize that DCIts learned the practically exact coefficient values used during data generation. Furthermore, for $X_{3,t}$, DCIts correctly identified that the generating process contains only linear cross-series lagged inputs from $X_{1,t-3}$ and $X_{2,t-5}$, with no cubic contribution to the third series. When examining the sign of the linear $\bm{\alpha}^{(1)}$ coefficients, we can see that DCIts learned that the linear term for the first two time series has a negative sign. We have the ground truth values $\alpha^{(1),\text{gt}}_{1,1,3} = \alpha^{(1),\text{gt}}_{2,2,5} = -2.75$. The DCIts results are $\alpha^{(1)}_{1,1,3} = (-2.749 \pm 0.04)$ and $\alpha^{(1)}_{2,2,5} = (-2.7501 \pm 0.0006)$. DCIts also correctly learned that there is no quadratic order present, i.e., all $\bm{\alpha}^{(2)}$ values are zero.

\section{Conclusion}\label{conclusion}

We presented the Deep Convolutional Interpreter for Time Series, an interpretable deep-learning architecture that exposes an explicit, sample-specific description of interaction structure in nonlinear multivariate time series, using forecasting as a training objective. The central object learned by the model is a lag-dependent transition tensor that maps recent observations to the next-step forecast. By factorizing this tensor into a sparse \emph{Focuser} and a signed \emph{Modeler}, DCIts produces, for every forecast and target series, a local lag-adjacency pattern together with signed coefficient values. In the higher-order extension, the same framework yields order-resolved elementwise polynomial contributions, enabling direct inspection of such nonlinear source--lag terms without resorting to post-hoc attribution.

Across benchmark datasets with known ground truth and additional controlled processes, DCIts achieved forecasting accuracy comparable to or better than the selected strong interpretable baseline, while recovering stable source--lag interaction patterns in the cases tested. In linear settings (including VAR-type dynamics), the learned coefficients accurately reconstruct the generating matrices and identify the active lags, strengths, and signs of interactions. In mixed regimes with bias and inhibitory couplings, the model retains sign fidelity and correctly resolves lag structure. In nonlinear settings, the polynomial extension recovers the correct interaction order and identifies cases in which higher-order terms are absent.

A key practical outcome is that interpretability is not an auxiliary diagnostic: it is produced \emph{in the course of prediction}. It remains local to each sample, allowing for the direct examination of regime-dependent interaction patterns. This supports the use of DCIts as a data-driven tool for probing effective connectivity and time-lagged coupling in complex dynamical systems, in addition to forecasting.

While we validate predictive competitiveness against IMV-LSTM on the established synthetic benchmark suite, we intentionally do not provide an exhaustive comparison against all major forecasting families (e.g., TCN/1D-CNN or all-MLP mixers). IMV-LSTM~\cite{guo2019exploring} was selected as the main comparator because it was identified in our previous benchmark as the strongest relevant interpretable deep baseline when forecasting performance and interpretability reliability were considered jointly on this dataset family~\cite{baric2021}. 

Our aim is to test whether intrinsic coefficient-level interpretability can be achieved without a substantial loss of forecasting accuracy in a directly comparable setting; the primary focus of follow-up work is therefore to extend and refine the interpretability analysis (e.g., robustness, stability, and regime dependence) rather than to build a broad forecasting leaderboard. Similarly, DCIts is not intended to replace classical VAR models when the data are well described by a low-order global linear process. In that regime, VAR is simpler and more efficient. The purpose of the VAR example in this paper is instead to verify that DCIts can recover stable signed coefficients in a setting where the correct transition matrices are known exactly, before moving to nonlinear and regime-dependent benchmark systems.

The current study also highlights limitations. The present higher-order formulation does not include mixed multiplicative terms across distinct source--lag entries, such as $X_{i,t-\ell_1}X_{j,t-\ell_2}$. It should therefore be interpreted as an elementwise polynomial expansion rather than as a complete multivariate polynomial basis. In the switching dataset, DCIts detected a regime-dependent interaction structure across multiple series but did not fully recover a regime-dependent bias shift in the switching driver series. This points to a concrete direction for refinement: extending the bias term to depend on lagged inputs (or adopting a gated/interacting bias mechanism) while preserving the model’s explicit coefficient structure. More broadly, higher-order expansions increase computational cost with window length and interaction order, motivating future work on structured sparsity, low-rank constraints across orders, and multi-scale temporal features.

Overall, within the controlled benchmark setting considered here, DCIts provides a forecasting framework that remains competitive in accuracy while yielding an explicit, lag-resolved, and locally interpretable representation of coupling structure, thereby helping to narrow the gap between high-capacity prediction and mechanism-oriented analysis in nonlinear multivariate time series.

\begin{acknowledgments}
D.H. acknowledges support from the project “Implementation of cutting-edge research and its application as part of the Scientific Center of Excellence for Quantum and Complex Systems, and Representations of Lie Algebras”, Grant No. PK.1.1.10.0004, co-financed by the European Union through the European Regional Development Fund - Competitiveness and Cohesion Programme 2021-2027, and Croatian Science Foundation (HRZZ) project IP-2022-10-1648. The authors thank Dario Bojanjac and Bartol Pavlović for useful discussions.
\end{acknowledgments}

\section*{Data Availability Statement}
Data openly available in a public repository that does not issue DOIs. All code, scripts, and Jupyter notebooks required to reproduce the experiments (Datasets 1--8) are publicly available in the DCIts GitHub repository \cite{githubdcits} \url{https://github.com/hc-xai/dcits}.

\appendix
\section{Benchmarking datasets}
\label{bdata}
The benchmarking datasets used in this study were sourced from Barić et al.~\cite{baric2021}. In Table~\ref{tab:benchmarking_datasets}, we provide detailed descriptions of all the datasets utilized in this work. The models used for benchmarking increase in complexity with each dataset, starting from a constant time series (Dataset 1) and progressing through various stages: autoregressive (Dataset 2), nonlinear autoregressive with no interactions between time series (Dataset 3), two interdependent time series  (Dataset 4), linear (Dataset 5), and nonlinear autoregressive (Dataset 6) with all subsequent time series derived from the first, custom vector autoregression model (Dataset 7), and a switching time series (Dataset 8). Dataset 8 was adjusted relative to the one presented in~\cite{baric2021}, with the first time series now incorporating the switching process to test the DCIts.

\begin{table*}[!ht]
\centering
\caption{Benchmarking datasets used in this study.}
\label{tab:benchmarking_datasets}
\begin{tabular}{lll}
\toprule
\textbf{Name} & \textbf{Model} & \textbf{Parameters} \\ \midrule
Dataset 1 &
$X_{n,t}=\alpha^{\mathrm{gt}}_{n} + \epsilon_{n,t}$ &
$\alpha^{\mathrm{gt}}_{n}\sim \mathcal{N}(0,1)$ \\ \midrule
Dataset 2 &
$X_{n,t}= \alpha^{\mathrm{gt}}_{n,n,l}X_{n,t-l}+ \epsilon_{n,t}$ &
$\alpha^{\mathrm{gt}}_{n,n,3}=\alpha^{\mathrm{gt}}_{n,n,7}=\tfrac{1}{2}$ \\ \midrule
Dataset 3 &
$X_{n,t}=\tanh\!\bigl(\alpha^{\mathrm{gt}}_{n,n,l}X_{n,t-l}\bigr)+\epsilon_{n,t}$ &
$\alpha^{\mathrm{gt}}_{n,n,3}=\tfrac{5}{7},\; \alpha^{\mathrm{gt}}_{n,n,7}=\alpha^{\mathrm{gt}}_{n,n,9}=\tfrac{1}{7}$ \\ \midrule
Dataset 4 & 
\begin{tabular}{@{}l@{}}
$\begin{array}{l}
X_{1,t} = \alpha^\text{gt}_{1,2,l} X_{2,t-l} + \epsilon_{1,t} \\
X_{2,t} = \alpha^\text{gt}_{2,1,l} X_{1,t-l} + \epsilon_{2,t}
\end{array}$
\end{tabular}& 
\begin{tabular}{@{}l@{}}
$\begin{array}{l}
\alpha^\text{gt}_{1,2,2}=\alpha^\text{gt}_{1,2,9}=\frac{2}{5}, \alpha^\text{gt}_{1,2,5}=\frac{1}{5} \\
\alpha^\text{gt}_{2,1,2}=\alpha^\text{gt}_{2,1,9}=\frac{2}{5}, \alpha^\text{gt}_{2,1,5}=\frac{1}{5} 
\end{array}$ 
\end{tabular} \\ 
\midrule
Dataset 5 &
$X_{n,t}=\alpha^\text{gt}_{n,1,l} X_{1,t-l}+ \epsilon_{n,t}$ &
\begin{tabular}{@{}l@{}}
$\renewcommand{\arraystretch}{1.1} 
\begin{array}{r@{=}l}
\alpha^\text{gt}_{1,1,3} & \frac{1}{2}, \alpha^\text{gt}_{1,1,4} = \frac{1}{2}, 
\alpha^\text{gt}_{2,1,9} = 1 \\
\alpha^\text{gt}_{3,1,2} & \frac{1}{2},  \alpha^\text{gt}_{3,1,7} = \frac{1}{2} \\
\alpha^\text{gt}_{4,1,3} & \frac{1}{10}, \alpha^\text{gt}_{4,1,4} = \frac{1}{10}, \alpha^\text{gt}_{4,1,8} = \frac{4}{5} \\
\alpha^\text{gt}_{5,1,2} & \frac{1}{3},  \alpha^\text{gt}_{5,1,5} = \frac{2}{9}, \alpha^\text{gt}_{5,1,8} = \frac{4}{9}
\end{array}$ 
\end{tabular} \\
\midrule
Dataset 6 &
$X_{n,t}=\tanh(\alpha^\text{gt}_{n,1,l} X_{1,t-l}) + \epsilon_{n,t}$ &
Identical to Dataset 5\\
\midrule
Dataset 7 &
\begin{tabular}{@{}l@{}}
$\begin{array}{l}
X_{1,t} = \alpha^\text{gt}_{1,1,1} X_{1,t-1} + \alpha^\text{gt}_{1,1,5} X_{1,t-5} + \epsilon_{1,t} \\
X_{2,t} = 1 + \alpha^\text{gt}_{2,1,2} X_{1,t-2} + \epsilon_{2,t} \\
X_{3,t} = \alpha^\text{gt}_{3,2,1} X_{2,t-1} + \alpha^\text{gt}_{3,4,4} X_{4,t-4} + \epsilon_{3,t} \\
X_{4,t} = 1 + \alpha^\text{gt}_{4,3,4} X_{3,t-4} + \alpha^\text{gt}_{4,5,1} X_{5,t-1} + \epsilon_{4,t} \\
X_{5,t} = \alpha^\text{gt}_{5,5,4} X_{5,t-4} + \alpha^\text{gt}_{5,2,1} X_{2,t-1} + \epsilon_{5,t}
\end{array}$ 
\end{tabular} &
\begin{tabular}{@{}l@{}}
$\renewcommand{\arraystretch}{1.1} 
\begin{array}{r@{=}l}
\alpha^\text{gt}_{1,1,1} & \frac{1}{4}, \alpha^\text{gt}_{1,1,5} = \frac{3}{4} \\
\alpha^\text{gt}_{2,1,2} & -1 \\
\alpha^\text{gt}_{3,2,1} & 1, \alpha^\text{gt}_{3,4,4} = 1 \\
\alpha^\text{gt}_{4,3,4} & -\frac{2}{7}, \alpha^\text{gt}_{4,5,1} = -\frac{5}{7} \\
\alpha^\text{gt}_{5,5,4} & \frac{12}{22}, \alpha^\text{gt}_{5,2,1} = \frac{10}{22}
\end{array}$ 
\end{tabular} \\
\midrule
Dataset 8 & 
\begin{tabular}{@{}l@{}}
$\begin{array}{l}
\text{if } X_{1,t-5} > \frac{1}{2}: \\
X_{1,t} =  b_\text{high} + \epsilon_{1,t} \\
X_{2,t} = \alpha^\text{gt}_{2,1,5} X_{1,t-5} + \epsilon_{2,t} \\
X_{3,t} = \alpha^\text{gt}_{3,1,4} X_{1,t-4} + \epsilon_{3,t} \\
X_{4,t} = \alpha^\text{gt}_{4,4,1} X_{4,t-1} + \alpha^\text{gt}_{4,4,4} X_{4,t-4} + \epsilon_{4,t} \\
\text{else:} \\
X_{1,t} = b_\text{low}  + \epsilon_{1,t} \\
X_{2,t} = \alpha^\text{gt}_{2,4,2} X_{4,t-2} + \epsilon_{2,t} \\
X_{3,t} = \alpha^\text{gt}_{3,4,4} X_{4,t-4} + \epsilon_{3,t} \\
X_{4,t} = \alpha^\text{gt}_{4,4,1} X_{4,t-1} + \alpha^\text{gt}_{4,4,4} X_{4,t-4} + \epsilon_{4,t} \\
\end{array}$ 
\end{tabular} & 
$\renewcommand{\arraystretch}{1.1} 
\begin{array}{l}
\alpha^\text{gt}_{2,1,5} = \frac{4}{5}, \alpha^\text{gt}_{2,4,2}= \frac{2}{3} \\
\alpha^\text{gt}_{3,1,4} = \frac{2}{3}, \alpha^\text{gt}_{3,4,4}= \frac{4}{5} \\
\alpha^\text{gt}_{4,4,1} = \frac{1}{2}, \alpha^\text{gt}_{4,4,4} = \frac{2}{5}\\
b_\text{high} = \frac{7}{10}, b_\text{low} = \frac{2}{10}\\
b_\text{high}\text{ and }b_\text{low}\text{ switch, based}\\ 
\text{on a persistence duration, P.}\\
P \sim \max(P_\text{min}, \mathcal{N}(\mu_P, \sigma_P))
\end{array}$  \\
\bottomrule
\end{tabular}
\end{table*}

\newpage
\bibliography{dcits}

@article{
chen2023tsmixer,
title={{TSM}ixer: An All-{MLP} Architecture for Time Series Forecast-ing},
author={Si-An Chen and Chun-Liang Li and Sercan O Arik and Nathanael Christian Yoder and Tomas Pfister},
journal={Transactions on Machine Learning Research},
issn={2835-8856},
year={2023},
url={https://openreview.net/forum?id=wbpxTuXgm0},
note={}
}

@inproceedings{assaf2019explainable,
  title={Explainable Deep Neural Networks for Multivariate Time Series Predictions.},
  author={Assaf, Roy and Schumann, Anika},
  booktitle={IJCAI},
  pages={6488--6490},
  year={2019}
}

@article{rudin2019stop,
  title={Stop explaining black box machine learning models for high stakes decisions and use interpretable models instead},
  author={Rudin, Cynthia},
  journal={Nature Machine Intelligence},
  volume={1},
  number={5},
  pages={206--215},
  year={2019},
  publisher={Nature Publishing Group}
}

@inproceedings{lundberg2017unified,
  title={A unified approach to interpreting model predictions},
  author={Lundberg, Scott M and Lee, Su-In},
  booktitle={Advances in neural information processing systems},
  pages={4765--4774},
  year={2017}
}

@article{Rungeetal19,
  title = {Inferring causation from time series with perspectives in Earth system sciences},
  author = {Runge, J. and Bathiany, S. and Bollt, E. and Camps-Valls, G. and Coumou, D. and Deyle, E. and Glymour, C. and Kretschmer, M. and Mahecha, M.D. and Munoz-Mari, J. and van Nes, E.H. and Peters, J. and Quax, R. and Reichstein, M. and Scheffer, M. and Sch{\"o}lkopf, B. and Spirtes, P. and Sugihara, G. and Sun, J. and Zhang, K. and Zscheischler, J.},
  journal = {Nature Communications},
  volume = {10},
  number = {2553},
  year = {2019}
}

@article{Runge_2019,
   title={Detecting and quantifying causal associations in large nonlinear time series datasets},
   volume={5},
   ISSN={2375-2548},
   url={http://dx.doi.org/10.1126/sciadv.aau4996},
   DOI={10.1126/sciadv.aau4996},
   number={11},
   journal={Science Advances},
   publisher={American Association for the Advancement of Science (AAAS)},
   author={Runge, Jakob and Nowack, Peer and Kretschmer, Marlene and Flaxman, Seth and Sejdinovic, Dino},
   year={2019},
   month={Nov},
   pages={eaau4996}
}

@article{Runge2018,
  doi = {10.1063/1.5025050},
  url = {https://doi.org/10.1063/1.5025050},
  year = {2018},
  month = {July},
  publisher = {{AIP} Publishing},
  volume = {28},
  number = {7},
  pages = {075310},
  author = {J. Runge},
  title = {Causal network reconstruction from time series: From theoretical assumptions to practical estimation},
  journal = {Chaos: An Interdisciplinary Journal of Nonlinear Science}
}

@article{fauvel2020performance,
  title={A Performance-Explainability Framework to Benchmark Machine Learning Methods: Application to Multivariate Time Series Classifiers},
  author={Fauvel, Kevin and Masson, V{\'e}ronique and Fromont, {\'E}lisa},
  journal={arXiv preprint arXiv:2005.14501},
  year={2020}
}

@article{baric2021,
AUTHOR = {Barić, Domjan and Fumić, Petar and Horvatić, Davor and Lipic, Tomislav},
TITLE = {Benchmarking Attention-Based Interpretability of Deep Learning in Multivariate Time Series Predictions},
JOURNAL = {Entropy},
VOLUME = {23},
YEAR = {2021},
NUMBER = {2},
ARTICLE-NUMBER = {143},
URL = {https://www.mdpi.com/1099-4300/23/2/143},
PubMedID = {33503822},
ISSN = {1099-4300},
}

@misc{guo2019exploring,
      title={Exploring Interpretable LSTM Neural Networks over Multi-Variable Data}, 
      author={Tian Guo and Tao Lin and Nino Antulov-Fantulin},
      year={2019},
      eprint={1905.12034},
      archivePrefix={arXiv},
      primaryClass={cs.LG}
}

@Article{Ozyegen2022,
author={Ozyegen, Ozan
and Ilic, Igor
and Cevik, Mucahit},
title={Evaluation of interpretability methods for multivariate time series forecasting},
journal={Applied Intelligence},
year={2022},
month={Mar},
day={01},
volume={52},
number={5},
pages={4727-4743},
issn={1573-7497},
doi={10.1007/s10489-021-02662-2},
url={https://doi.org/10.1007/s10489-021-02662-2}
}

@article{perturbation,
  author       = {Udo Schlegel and
                  Daniela Oelke and
                  Daniel A. Keim and
                  Mennatallah El{-}Assady},
  title        = {An Empirical Study of Explainable {AI} Techniques on Deep Learning
                  Models For Time Series Tasks},
  journal      = {CoRR},
  volume       = {abs/2012.04344},
  year         = {2020},
  url          = {https://arxiv.org/abs/2012.04344},
  eprinttype    = {arXiv},
  eprint       = {2012.04344},
  bibsource    = {dblp computer science bibliography, https://dblp.org}
}

@article{GANDIN2021103876,
title = {Interpretability of time-series deep learning models: A study in cardiovascular patients admitted to Intensive care unit},
journal = {Journal of Biomedical Informatics},
volume = {121},
pages = {103876},
year = {2021},
issn = {1532-0464},
doi = {https://doi.org/10.1016/j.jbi.2021.103876},
url = {https://www.sciencedirect.com/science/article/pii/S1532046421002057},
author = {Ilaria Gandin and Arjuna Scagnetto and Simona Romani and Giulia Barbati}
}

@misc{Feng2022,
      title={Multi-scale Attention Flow for Probabilistic Time Series Forecasting}, 
      author={Shibo Feng and Chunyan Miao and Ke Xu and Jiaxiang Wu and Pengcheng Wu and Yang Zhang and Peilin Zhao},
      year={2023},
      eprint={2205.07493},
      archivePrefix={arXiv},
      primaryClass={cs.LG}
}

@INPROCEEDINGS{9308570, author={Pantiskas, Leonardos and Verstoep, Kees and Bal, Henri}, booktitle={2020 IEEE Symposium Series on Computational Intelligence (SSCI)}, title={Interpretable Multivariate Time Series Forecasting with Temporal Attention Convolutional Neural Networks}, year={2020}, volume={}, number={}, pages={1687-1694}, doi={10.1109/SSCI47803.2020.9308570}}

@Article{Wu2023,
author={Wu, Binrong
and Wang, Lin
and Zeng, Yu-Rong},
title={Interpretable tourism demand forecasting with temporal fusion transformers amid COVID-19},
journal={Applied Intelligence},
year={2023},
month={Jun},
day={01},
volume={53},
number={11},
pages={14493-14514},
issn={1573-7497},
doi={10.1007/s10489-022-04254-0},
url={https://doi.org/10.1007/s10489-022-04254-0}
}

@InProceedings{Pham2023,
author="Pham, Anh-Duy
and Kuestenmacher, Anastassia
and Ploeger, Paul G.",
editor="Arai, Kohei",
title="TSEM: Temporally-Weighted Spatiotemporal Explainable Neural Network for Multivariate Time Series",
booktitle="Advances in Information and Communication",
year="2023",
publisher="Springer Nature Switzerland",
address="Cham",
pages="183--204",
isbn="978-3-031-28073-3"
}

@article{Tryambak2020,
  author       = {Tryambak Gangopadhyay and
                  Sin Yong Tan and
                  Zhanhong Jiang and
                  Rui Meng and
                  Soumik Sarkar},
  title        = {Spatiotemporal Attention for Multivariate Time Series Prediction and
                  Interpretation},
  journal      = {CoRR},
  volume       = {abs/2008.04882},
  year         = {2020},
  url          = {https://arxiv.org/abs/2008.04882},
  eprinttype    = {arXiv},
  eprint       = {2008.04882},
  bibsource    = {dblp computer science bibliography, https://dblp.org}
}

@article{attentionNot,
  author       = {Sarthak Jain and
                  Byron C. Wallace},
  title        = {Attention is not Explanation},
  journal      = {CoRR},
  volume       = {abs/1902.10186},
  year         = {2019},
  url          = {http://arxiv.org/abs/1902.10186},
  eprinttype    = {arXiv},
  eprint       = {1902.10186},
  bibsource    = {dblp computer science bibliography, https://dblp.org}
}

@article{attentionNotNot,
  author       = {Sarah Wiegreffe and
                  Yuval Pinter},
  title        = {Attention is not not Explanation},
  journal      = {CoRR},
  volume       = {abs/1908.04626},
  year         = {2019},
  url          = {http://arxiv.org/abs/1908.04626},
  eprinttype    = {arXiv},
  eprint       = {1908.04626},
  bibsource    = {dblp computer science bibliography, https://dblp.org}
}

@article{Oreshkin2019,
  author       = {Boris N. Oreshkin and
                  Dmitri Carpov and
                  Nicolas Chapados and
                  Yoshua Bengio},
  title        = {{N-BEATS:} Neural basis expansion analysis for interpretable time
                  series forecasting},
  journal      = {CoRR},
  volume       = {abs/1905.10437},
  year         = {2019},
  url          = {http://arxiv.org/abs/1905.10437},
  eprinttype    = {arXiv},
  eprint       = {1905.10437},
  bibsource    = {dblp computer science bibliography, https://dblp.org}
}

@article{Rasul2020,
  author       = {Kashif Rasul and
                  Abdul{-}Saboor Sheikh and
                  Ingmar Schuster and
                  Urs Bergmann and
                  Roland Vollgraf},
  title        = {Multi-variate Probabilistic Time Series Forecasting via Conditioned
                  Normalizing Flows},
  journal      = {CoRR},
  volume       = {abs/2002.06103},
  year         = {2020},
  url          = {https://arxiv.org/abs/2002.06103},
  eprinttype    = {arXiv},
  eprint       = {2002.06103},
  bibsource    = {dblp computer science bibliography, https://dblp.org}
}

@article{Wang2018,
  author       = {Jingyuan Wang and
                  Ze Wang and
                  Jianfeng Li and
                  Junjie Wu},
  title        = {Multilevel Wavelet Decomposition Network for Interpretable Time Series
                  Analysis},
  journal      = {CoRR},
  volume       = {abs/1806.08946},
  year         = {2018},
  url          = {http://arxiv.org/abs/1806.08946},
  eprinttype    = {arXiv},
  eprint       = {1806.08946},
  bibsource    = {dblp computer science bibliography, https://dblp.org}
}

@article{DualStage,
  author       = {Yao Qin and
                  Dongjin Song and
                  Haifeng Chen and
                  Wei Cheng and
                  Guofei Jiang and
                  Garrison W. Cottrell},
  title        = {A Dual-Stage Attention-Based Recurrent Neural Network for Time Series
                  Prediction},
  journal      = {CoRR},
  volume       = {abs/1704.02971},
  year         = {2017},
  url          = {http://arxiv.org/abs/1704.02971},
  eprinttype    = {arXiv},
  eprint       = {1704.02971},
  bibsource    = {dblp computer science bibliography, https://dblp.org}
}

@article{castro2023time,
  title = {Time series causal relationships discovery through feature importance and ensemble models},
  volume = {13},
  ISSN = {2045-2322},
  url = {http://dx.doi.org/10.1038/s41598-023-37929-w},
  DOI = {10.1038/s41598-023-37929-w},
  number = {1},
  journal = {Scientific Reports},
  publisher = {Springer Science and Business Media LLC},
  author = {Castro,  Manuel and Mendes Júnior,  Pedro Ribeiro and Soriano-Vargas,  Aurea and de Oliveira Werneck,  Rafael and Moreira Gon\c{c}alves,  Maiara and Lusquino Filho,  Leopoldo and Moura,  Renato and Zampieri,  Marcelo and Linares,  Oscar and Ferreira,  Vitor and Ferreira,  Alexandre and Davólio,  Alessandra and Schiozer,  Denis and Rocha,  Anderson},
  year = {2023},
  month = jul 
}

@misc{githubdcits,
  author       = {D. Barić and D. Horvatić},
  title        = {{DCIts}},
  year         = {2024},
  howpublished = {\url{https://github.com/hc-xai/dcits}},
  note         = {Maintained by the authors, Accessed: January 26th 2026}
}

@misc{vansprang2024conceptbottleneck,
     title={Interpretability for Time Series Transformers using A Concept Bottleneck Framework}, 
      author={Angela van Sprang and Erman Acar and Willem Zuidema},
      year={2025},
      eprint={2410.06070},
      archivePrefix={arXiv},
      primaryClass={cs.LG},
      url={https://arxiv.org/abs/2410.06070}, 
}

@misc{xforecast2024,
      title={XForecast: Evaluating Natural Language Explanations for Time Series Forecasting}, 
      author={Taha Aksu and Chenghao Liu and Amrita Saha and Sarah Tan and Caiming Xiong and Doyen Sahoo},
      year={2024},
      eprint={2410.14180},
      archivePrefix={arXiv},
      primaryClass={cs.CL},
      url={https://arxiv.org/abs/2410.14180}, 
}

@inproceedings{wang_counterfactual_2023,
  author={Wang, Zhendong and Miliou, Ioanna and Samsten, Isak and Papapetrou, Panagiotis},
  booktitle={2023 IEEE International Conference on Data Mining (ICDM)}, 
  title={Counterfactual Explanations for Time Series Forecasting}, 
  year={2023},
  volume={},
  number={},
  pages={1391-1396},
  doi={10.1109/ICDM58522.2023.00180}}

@article{lim2021time,
  title={Time-series forecasting with deep learning: a survey},
  author={Lim, Bryan and Zohren, Stefan},
  journal={Philosophical Transactions of the Royal Society A},
  volume={379},
  number={2194},
  pages={20200209},
  year={2021},
  publisher={The Royal Society}
}

@article{benidis2022deep,
  title={Deep learning for time series forecasting: Tutorial and literature survey},
  author={Benidis, Konstantinos and Rangapuram, Syama Sundar and Flunkert, Valentin and Wang, Yuyang and Maddix, Danielle and Turkmen, Caner and Gasthaus, Jan and Bohlke-Schneider, Michael and Salinas, David and Stella, Lorenzo and others},
  journal={ACM Computing Surveys},
  volume={55},
  number={6},
  pages={1--36},
  year={2022},
  publisher={ACM}
}

@article{tonekaboni2020went,
  title={What went wrong and when? Instance-wise feature importance for time-series black-box models},
  author={Tonekaboni, Sana and Joshi, Shalmali and Campbell, Kieran and Duvenaud, David K and Goldenberg, Anna},
  journal={Advances in Neural Information Processing Systems},
  volume={33},
  pages={799--809},
  year={2020}
}

\end{document}